\documentclass[nonblindrev]{informs3}

\DoubleSpacedXI 

\usepackage{endnotes}
\let\footnote=\endnote
\let\enotesize=\normalsize
\def\notesname{Endnotes}%
\def\makeenmark{\hbox to1.275em{\theenmark.\enskip\hss}}
\def\enoteformat{\rightskip0pt\leftskip0pt\parindent=1.275em
  \leavevmode\llap{\makeenmark}}

\usepackage{rotating}
\usepackage{fancyvrb}
\usepackage{algorithm}
\usepackage{algpseudocode}
\usepackage{tikz}
\usepackage{diagbox}
\usepackage{mathrsfs}
\usepackage{booktabs} 
\usepackage{longtable}
\usepackage{amsmath}
\usepackage{amssymb}
\usepackage{newtxtext,newtxmath}
\usepackage{algorithm}
\usepackage{algpseudocode}
\usepackage{setspace}
\usepackage{anyfontsize}
\usepackage{natbib}
 \bibpunct[, ]{(}{)}{,}{a}{}{,}%
 \def\bibfont{\small}%
\TheoremsNumberedThrough     
\ECRepeatTheorems

\EquationsNumberedThrough    

\begin{document}


\RUNAUTHOR{Wang et al.}

\RUNTITLE{DRO-LS}

\TITLE{Generalizing Few Data to Unseen Domains Flexibly Based on Label Smoothing Integrated with Distributionally Robust Optimization}

\ARTICLEAUTHORS{%
\AUTHOR{Yangdi Wang}
\AFF{School of Management, Beijing Institute of Technology, Beijing, 100081, \EMAIL{wangyangdi@bit.edu.cn}} 
\AUTHOR{Zhi-Hai Zhang$^\dag$}
\AFF{Department of Industrial Engineering, Tsinghua University, Beijing, China, 100084, \EMAIL{zhzhang@tsinghua.edu.cn}} 
\AUTHOR{Su Xiu Xu$^\dag$}
\AFF{School of Management, Beijing Institute of Technology, Beijing, China, 100081, \EMAIL{xusuxiu@bit.edu.cn}}
\AUTHOR{Wenming Guo}
\AFF{School of Computer Science (National Pilot Software Engineering School), Beijing University of Posts and Telecommunications, Beijing, China, 100876, \EMAIL{guowenming@bupt.edu.cn}}
} 

\ABSTRACT{%
Overfitting commonly occurs when applying deep neural networks (DNNs) on small-scale datasets, where DNNs do not generalize well from existing data to unseen data. The main reason resulting in overfitting is that small-scale datasets cannot reflect the situations of the real world. Label smoothing (LS) is an effective regularization method to prevent overfitting, avoiding it by mixing one-hot labels with uniform label vectors. However, LS only focuses on labels while ignoring the distribution of existing data. In this paper, we introduce the distributionally robust optimization (DRO) to LS, achieving shift the existing data distribution flexibly to unseen domains when training DNNs. Specifically, we prove that the regularization of LS can be extended to a regularization term for the DNNs parameters when integrating DRO. The regularization term can be utilized to shift existing data to unseen domains and generate new data. Furthermore, we propose an approximate gradient-iteration label smoothing algorithm (GI-LS) to achieve the findings and train DNNs. We prove that the shift for the existing data does not influence the convergence of GI-LS. Since GI-LS incorporates a series of hyperparameters, we further consider using Bayesian optimization (BO) to find the relatively optimal combinations of these hyperparameters. Taking small-scale anomaly classification tasks as a case, we evaluate GI-LS, and the results clearly demonstrate its superior performance.
}%



\maketitle

%

\section{Introduction}\label{sec:1}\
In supervised deep learning, overfitting is a common issue when training machine learning deep neural networks (DNNs), where DNNs perform well on training data but struggle to generalize to unseen data. This issue is especially frequent when the dataset is small, as a small-scale dataset cannot accurately reflect real-world conditions (\citeauthor{zhou2022domain}, \citeyear{zhou2022domain}). Due to real-world constraints (e.g., technical difficulty, inclusion of private information, etc.), it is unrealistic to collect large-scale data in some industries. For instance, the anomaly data collection in the process of producing various industrial products is always extremely time-consuming and expensive. Thus, there is a scarcity of anomaly data when maintaining and controlling the quality of products (\citeauthor{bergmann2019mvtec}, \citeyear{bergmann2019mvtec}; \citeauthor{tabernik2020segmentation}, \citeyear{tabernik2020segmentation}). This dilemma is also present in other domains such as medical image analysis, financial development trend prediction and so on. A suitable approach to preventing overfitting while improving the generalization of DNNs using existing data could lead to advancements in a wide range of fields.
\par In supervised deep learning, the generalization of DNNs can be improved from both the model itself and the dataset information perspectives. Recently, a variety of regularization strategies have emerged to prevent overfitting and improve model's generalization. One branch of regularization strategies, such as \(l_1\) or \(l_2\) penalization, weight decay, and dropout, constrains the range of variation for model parameters or acts on hidden activation functions during model training. However, these methods intricately involve the specific parameterization or loss criterion of the model (\citeauthor{li2020regularization}, \citeyear{li2020regularization}), complicating the training process. In contrast, regularization strategies that prevent overfitting from the viewpoint of the dataset do not influence the structure of the model, making the training process simpler. The most direct method is data augmentation expanding the scale of dataset. However, collecting and annotating data in some industries always incur extremely high costs, are time-consuming, and laborious in many industries (\citeauthor{aghasi2024deep}, \citeyear{aghasi2024deep}). It is natural to use the existing data to generate new ones. Unfortunately, the generated data might change the existing data characteristics significantly, potentially leading to poor performance of model on unseen data (\citeauthor{peck2023introduction}, \citeyear{peck2023introduction}). Beyond merely considering existing data, since each data point in supervised deep learning contains labels with annotation information (e.g., category information) that are used to fit the output distribution when training DNNs, some regularization strategies directly utilize these labels to prevent overfitting. A typical regularization method is label smoothing (LS), which contains only one hyperparameter for labels and shows superior effectiveness in preventing overfitting when training DNNs (\citeauthor{szegedy2016rethinking}, \citeyear{szegedy2016rethinking}; \citeauthor{muller2019does}, \citeyear{muller2019does}).
\subsection{Short Introduction to Label Smoothing}\label{sec1.1}\
In this section, we briefly introduce LS using the \(k\)-class supervised classification task.
\par Taking arbitrary sample \((X, Y)\) belonging to a certain category $i$ from dataset, we consider the hard one-hot vector to represent its corresponding label $Y$ when training DNNs.
\begin{equation}
Y = \boldsymbol[y_{1}, \quad y_{2}, \quad...\quad, \quad y_{k}\boldsymbol],
\label{2} 
\end{equation}
where $y_{i} \in Y$ is:
\begin{equation}
y_i=\left\{
\begin{aligned}
&1, \qquad& &\text{if sample \textit{X} belongs to class \textit{i},} \\
&0  \qquad& \  & \text{otherwise.}
\end{aligned}
\label{3}
\right.
\end{equation}
\par In general, the output of model may closely mirror the hard one-hot label $Y$, where the prediction for the \(i^{th}\) position approaches $1$, while the predictions for other positions are nearly $0$. This issue tends to result in the overconfidence of model's output and leads to overfitting.
Furthermore, the situation is particularly evident when the scale of dataset is small. To address this issue, LS considers changing the hard one-hot label $Y$ in the following.
\begin{equation}
y_i=\left\{
\begin{aligned}
&1-\alpha, \qquad& &\text{if sample \textit{X} belongs to class \textit{i},} \\
&\frac{\alpha}{k-1} \qquad& \  & \text{otherwise.}
\end{aligned}
\label{5}
\right.
\end{equation}
where $\alpha$ is a hyperparameter adjusted manually.
\par LS only introduces a hyperparameter \(\alpha\) to adjust each position of the hard one-hot label \(Y\). When $\alpha = 0$, LS degenerates to hard one-hot label in equation (\ref{2}). As $\alpha$ increases, the \(i^{th}\) position changes from 1 to \(1-\alpha\) and the other \(k-1\) positions change from 0 to \(\frac{\alpha}{k-1}\). In this process, the true label value progressively decreases while the values of other categories gradually increase, penalizing the overconfidence in the outputs for the true categories while giving some attention to other categories. This makes the output distribution relatively smoother, thus improving the model's generalization. With the variation of $\alpha$, we actually assign multiple labels to the same data without extra annotations. 
\par There exists a body of literature providing explanations regarding the superior regularization effect of LS from both theoretical and experimental perspectives and further improving it (\citeauthor{muller2019does}, \citeyear{muller2019does}; \citeauthor{lukasik2020does}, \citeyear{lukasik2020does}; \citeauthor{yuan2020revisiting}, \citeyear{yuan2020revisiting}). 
However, these research does not break the dependence of LS on labels. There is still a lack of research on expanding its regularization effect by overcoming the limitations imposed by labels.
\par Due to the restriction of the dataset's scale, one challenge is how to shift existing data to appropriate unseen domains, allowing DNNs to learn more effective characteristic information about the data and prevent overfitting. It is well-known that the distributionally robust optimization (DRO) can build relationship between robustness and regularization under certain conditions (\citeauthor{shafieezadeh2015distributionally}, \citeyear{shafieezadeh2015distributionally}; \citeauthor{kuhn2019Wasserstei}, \citeyear{kuhn2019Wasserstei}). \cite{sinha2018certifiable} and \cite{bai2024wasserstein} present that the robustness equals to a regularization term, which can be utilized to shift the existing data to unseen domains and generate new data. One strength of DRO is that it can ensure the generated data is closely related to the existing data, thereby maintaining the original characteristics to some extent. However, DRO only considers the distribution in the worst-case scenario (\citeauthor{jiao2022distributed}, \citeyear{jiao2022distributed}), which limits its flexibility to shift data to unseen data further. 
\subsection{Our Contributions}\label{sec:1.2}\
\par In this paper, based on the powerful regularization effect of LS for labels and its hyperparameter when training DNNs, we integrate LS into the regularization term built by DRO, introducing multiple worst-case scenarios and enabling the exploration of appropriate data shifts, adding flexibility to DRO. These data shifts are equivalent to generating multiple new samples for the same label without extra annotations. Furthermore, we overcome the limitations imposed by labels on LS by incorporating DRO. 
\par Our main contribution are summarized as follows. 
\begin{itemize}
    \item We propose a novel two-stage problem, named DRO-LS, integrated LS within the DRO framework for generalizing few data to unseen domains flexibly. Specifically, the first stage perturbs existing data by LS and generate new samples, while the second stage involves using the original data and generated data to train DNNs. 
    \item To the best of our knowledge, it is the first time that the regularization effect of LS overcomes the limitations imposed by labels and has been applied to generate data. In addition, it is the first time that the worst-case scenario of DRO can be adjusted flexibly. Specifically, the regularization term built by DRO roughly corresponds to an \(L_2\) penalization that integrates the regularization effect of LS and DNNs parameters. Moreover, we prove that there exists a bound between the existing data and the generated data, showing that the generated data maintains the characteristics of the existing data even if the hyperparameter of LS varies.
    \item We develop a two-stage gradient-based algorithm, called gradient-iteration label smoothing (GI-LS), to approximately solve DRO-LS model. The algorithm can be regarded as a new data augmentation algorithm. Furthermore, we prove the convergence of the proposed algorithm, illustrating that perturbations on existing data with LS do not affect its convergence rate.
    \item Since the proposed algorithm involves a series of hyperparameters for adjustment, we take in specific small-scale anomaly cases to conduct extensive simulations using Bayesian optimization (BO) provide suggestions for selecting the hyperparameter ranges of GI-LS. Furthermore, as a perturbation algorithm for existing data, we validate the effectiveness of GI-LS in defending against some general adversarial attack methods in the context of magnetic surface defects.
\end{itemize}
\par The rest of this paper is organized as follows. Section \ref{sec:2} provides the relevant literature. Section \ref{sec:3} introduces the base model and algorithm. In Section \ref{sec:4}, using anomaly classification as a case study, the algorithm performance is evaluated with simulation analyses using BO. Section \ref{sec:5} concludes the paper and presents future work. All proofs and experiments are provided in the Appendix.
\section{Related work}\label{sec:2}\
Our work is closely related to two streams of research in the literature. The first stream concerns data augmentation. It is common that the distribution of training data does not reflect the real-world situation, directly resulting in the overfitting of models. Data augmentation deals with the overfitting from the root of problem, which expands the scale of training data (\citeauthor{wang2021regularizing}, \citeyear{wang2021regularizing}). However, constrained by real-world conditions, collecting and annotating samples in some industries always incurs extremely high costs. Thus, designing data augmentation based on the existing data become the only choice. Some research achieves this by directly splicing, reassembling, and occluding data (\citeauthor{devries2017improved}, \citeyear{devries2017improved}; \citeauthor{zhang2018mixup}, \citeyear{zhang2018mixup}; \citeauthor{yun2019cutmix},\citeyear{yun2019cutmix}). Unfortunately, these methods might change the characteristics of original data significantly, potentially leading to poor generalization of DNNs on unseen data. In contrast, our work guarantees that the generated samples surround the existing data by introducing DRO. Additionally, there are another data augmentation methods in the literature that generates adversarial samples, based on point-wise \( l_p \)-bounded perturbations, to shift existing data (\citeauthor{luo2020data}, \citeyear{luo2020data}; \citeauthor{bai2024wasserstein}, \citeyear{bai2024wasserstein}). Typical adversarial methods include Fast Gradient Sign Method (FGSM) (\citeauthor{goodfellow2014explaining}, \citeyear{goodfellow2014explaining}), CW attacking (\citeauthor{carlini2017towards}, \citeyear{carlini2017towards}), Projected Gradient Descent (PGD) (\citeauthor{madry2018towards}, \citeyear{madry2018towards}), etc. See \cite{peck2023introduction} for a comprehensive review of adversarial data augmentation method. In general, these methods are often summarized as white-box attacks and cause a negative impact on DNNs performance (\citeauthor{bai2024wasserstein}, \citeyear{bai2024wasserstein}). It has been proven that DNNs are vulnerable to adversarial data augmentation (\citeauthor{goodfellow2014explaining}, \citeyear{goodfellow2014explaining}), where imperceptible adversarial perturbations significantly decreases the performance of DNNs (\citeauthor{peck2023introduction}, \citeyear{peck2023introduction}). The data augmentation proposed in our paper can also be regarded as a perturbation method. However, different from attacking DNNs with imperceptible perturbations, our work primarily focuses on giving large perturbations to flexibly shift data for small-scale datasets (\citeauthor{volpi2018generalizing}, \citeyear{volpi2018generalizing}).
\par The second related stream of literature concentrates on distributionally robustness optimization (DRO). DRO has been widely used in a wild range of field, such as machine scheduling, manufacturing, etc (\citeauthor{shu2014dynamic}, \citeyear{shu2014dynamic}; \citeauthor{noyan2022distributionally}, \citeyear{noyan2022distributionally}). In recent years, DRO has made great progress in building the relationship between robustness and regularization in machine learning (\citeauthor{kuhn2019Wasserstei}, \citeyear{kuhn2019Wasserstei}; \citeauthor{duchi2021statistics}, \citeyear{duchi2021statistics}; \citeauthor{bai2024wasserstein}, \citeyear{bai2024wasserstein}), making contribution to both convex and non-convex models (\citeauthor{shafieezadeh2019regularization}, \citeyear{shafieezadeh2019regularization}; \citeauthor{levy2020large}, \citeyear{levy2020large}). The crucial considerations for DRO in machine learning are how to choose the ambiguity set while maintain tractability. At present, most research considers using the Wasserstein distance, a metric for transforming one distribution into another, to build the ambiguity set (\citeauthor{blanchet2021statistical}, \citeyear{blanchet2021statistical}); \citeauthor{liu2022distributionally}, \citeyear{liu2022distributionally}). This method, called Wasserstein distributionally robust optimization (WDRO), minimizes the worst-case distribution within a Wasserstein ball, offering a new interpretation of the relationship between robustness and regularization (\citeauthor{kuhn2019Wasserstei}, \citeyear{kuhn2019Wasserstei}; \citeauthor{rychener2023end}, \citeyear{rychener2023end}). To guarantee tractability and further convert WDRO, Lagrangian relaxation and duality are often considered (\citeauthor{sinha2018certifiable}, \citeyear{sinha2018certifiable}; \citeauthor{blanchet2019quantifying}, \citeyear{blanchet2019quantifying}). \cite{shafieezadeh2015distributionally} developed a linear WDRO distributionally robust logistic regression model, which can be converted to a complementary regularization term related to the label using Lagrangian relaxation and duality. Furthermore, they prove that such a logistic regression model is equivalent to a tractable convex program. \cite{shafieezadeh2019regularization} integrated DNNs into WDRO. By considering Lagrangian relaxation and duality, they demonstrated that the optimization objective of the model can be approximated as an empirical risk minimization (ERM) problem with a convex regularization term by incorporating Lipschitz moduli. This program can be effectively addressed with a gradient-based method. Other gradient-based schemes for WDRO with DNNs can refer to \cite{garcia2022regularized}. Our work is similar to the WDRO frameworks proposed by \cite{sinha2018certifiable} and \cite{volpi2018generalizing}, but we consider the regularization effects of LS to further shift the worst-case distribution of DRO. \cite{sinha2018certifiable} points out that the samples generated by WDRO can defend against adversarial data augmentation methods. In our work, we select some adversarial data augmentation methods to validate GI-LS. We find that the defense mechanism remains valid even when we incorporate the regularization effect of LS.
\section{Method}\label{sec:3}\
\subsection{Problem formulation}\
In supervised learning, the optimization objective of training DNNs is as follows.
\begin{equation}
\mathop{\operatorname{min}}_{\theta \in \Theta}{E}_{ (X,Y) \sim P_{0}}[\ell(\theta ; (X, Y))],
\label{1}
\end{equation}
where $(X, Y) \in (\mathcal{X}, \mathcal{Y})$ are drawn from the training distribution $P_0$. $X$ denotes the training sample and $Y$ is the corresponding label. $\theta \in \Theta$ represents the parameters of DNNs. $\ell: \mathcal{X} \times \mathcal{Y} \rightarrow \mathbb{R}$ is the loss function.
\par We introduce the cross-entropy loss as the loss function $\ell$.
\begin{equation}
\ell(\theta ; (X, Y)) = -\sum_{i=1}^{k}\sum_{j=1}^{k}y_{i}\log p_{j}(\theta, x).
\label{4}
\end{equation}
\par Given the parameters of DNNs as \(\theta := (\theta_f, \theta_r)\), where \(\theta_r\) are the parameters of the hidden layers, and \(\theta_f \in \mathbb{R}^{t \times k}\) are the parameters of the final classification layer. Let \(m \times t\) be the dimension of the feature map output from the hidden layers. \(p_j(\theta, x)\) is given as follows.
\begin{equation*}
  p_j(\theta, x) := \frac{\exp \left(\theta_{f, j}^{\top} f(\theta_{r}, x)\right)}{\sum_{l=1}^{k} \exp \left(\theta_{f, l}^{\top} f\left(\theta_{r} ; x\right)\right),} 
\end{equation*}
where $\theta_{f, l}$ is the $l$-th column of classification layer weight $\theta_{f}$, and $f\left(\theta_{r} ; x\right) \in \mathbb{R}^{m \times t}$ is the feature map of the training data.
\par Specifically, we denote the cross entropy loss integrated LS as $\boldsymbol{\ell}$. 
\par Now, we define the Wasserstein distance in the following definition.
\begin{definition}[Wasserstein distance, \citeauthor{sinha2018certifiable}, \citeyear{sinha2018certifiable}; \citeauthor{villani2021topics}, \citeyear{villani2021topics}]
Let transport metric \( c: \mathcal{Z} \times \mathcal{Z} \rightarrow [0, \infty) \) be non-negative, lower semi-continuous, and satisfy \( c(z, z) = 0 \) for a probability space \((\mathcal{Z}, \mathcal{A}, P_0)\), where \( \mathcal{Z} \subseteq \mathbb{R}^{m}\). For probability measures \( P \) and \( Q \) supported on \( \mathcal{Z} \), \( \Pi(P, Q) \) denotes their couplings, which means measures \( M \) on \( \mathcal{Z}^{2} \) with \( M(A, \mathcal{Z}) = P(A) \) and \( M(\mathcal{Z}, A) = Q(A) \). The Wasserstein distance between $P$ and $Q$ is given as follows.
\begin{equation}
W_{c}(P, Q):= \inf _{M \in \Pi(P, Q)}\operatorname{E}_{M}\left[c\left(Z, Z^{\prime}\right)\right].
\label{6}
\end{equation}
\label{d1}
\end{definition}
\par The distance in the space of learned representations of the high-capacity model typically corresponds to the semantic distance in visual space (\citeauthor{volpi2018generalizing}, \citeyear{volpi2018generalizing}). We follow this setting and further define the Wasserstein distance on the semantic space.  
\par For the training sample \((x, y) \in (X, Y)\), we consider the transport metric \(c\) on the feature map \(f(\theta_r; x)\)
\begin{equation}
c_\theta\left((x, y),\left(x^{\prime}, y^{\prime}\right)\right) : = \frac{1}{2}\left\|f(\theta_{r} ; x) - f\left(\theta_{r} ; x^{\prime}\right) \right\|_2^2 + \mathcal{K}\|y - y^\prime\|,
\label{7}
\end{equation}
\par where $c_\theta(\cdot)$ represents the cost of moving mass from point $(f\left(\theta_{r} ; x\right), y)$ to $(f\left(\theta_{r} ; x^\prime\right), y^\prime)$. Since we primarily focus on the perturbation of the feature map \( f(\theta_r; x') \), we set \(\mathcal{K}\) to \(\infty\) if data points have different labels (\(y \neq y'\)). The Wasserstein distance on the semantic space is given as follows.
\begin{equation}
W_{\theta}(P, Q):=\inf _{M \in \Pi(P, Q)} \operatorname{E}_{M}\left[c_{\theta}\left((X, Y),\left(X^{\prime}, Y^{\prime}\right)\right)\right].
\label{8}
\end{equation}
\par For \(\rho \geq 0\), we define Wasserstein distribution ball: \(\mathcal{P}=\left\{P: W_\theta\left(P, P_0\right) \leq \rho\right\}\) and select the distribution \(P \in \mathcal{P}\) around the training distribution \(P_0\). Furthermore, we formulate the following problem:
\begin{equation}
\mathop{\operatorname{min}}_{\theta \in \Theta} \sup _{P \in \mathcal{P}} \operatorname{E}_{(X,Y)\sim P}[\boldsymbol\ell(\theta; (X, Y))].
\label{9}
\end{equation}
\par Unfortunately, the above supremum over probability distribution is always intractable (\citeauthor{blanchet2019quantifying}, \citeyear{blanchet2019quantifying}; \citeauthor{sinha2018certifiable}, \citeyear{sinha2018certifiable}; \citeauthor{gao2017distributional}, \citeyear{gao2017distributional}; \citeauthor{volpi2018generalizing}, \citeyear{volpi2018generalizing}). To solve this issue, we consider Lagrangian relaxation with a fixed penalty parameter \(\gamma \geq 0\) to reformulate the problem (\ref{9}).
\begin{equation}
    \min _{\theta \in \Theta} \sup _{P \in \mathcal{P}}\left\{\mathrm{E}_{(X, Y) \sim P}[\boldsymbol\ell(\theta ;(X, Y))]-\gamma W_\theta\left(P, P_0\right)]\right\}.
    \label{10}
\end{equation}
\par Let \(\mathcal{Z} \subseteq \mathbb{R}^{m \times t}\). We define the robust surrogate loss as follows.
\begin{equation}
    \boldsymbol{\phi}_{\boldsymbol{\gamma}}(\theta ; (z^\prime, y)) := \sup_{z \in \mathcal{Z}} \left\{\boldsymbol{\ell}(\theta ; (z, y)) - \frac{\gamma}{2} \|z - z'\|_2^2\right\} \text{,}
    \label{101}
\end{equation}
where $z^\prime = f(\theta_r; x)$. 
\par The following theorem illustrates the duality results of the supremum stage of the problem (\ref{10}) in the semantic space. 
\begin{theorem}
For any distribution $Q$ and any $\rho > 0$,
\begin{equation*}
\sup _{P: W_\theta(P, Q) \leq \rho} E_{(Z,Y) \sim P}[\boldsymbol\ell(\theta ; (Z, Y))]=\inf _{\gamma \geq 0}\left\{\gamma \rho+E_Q\left[\boldsymbol\phi_{\boldsymbol\gamma}(\theta; (Z^\prime, Y))\right]\right\}.
\end{equation*}
\par For any $\gamma \geq 0$, we have
\begin{equation}
\sup _{P: W_\theta(P, Q) \leq \rho}\left\{E_{(Z, Y) \sim P}[\boldsymbol\ell(\theta ; (Z, Y))] - \gamma W_\theta(P, Q)\right\}=E_Q\left[\boldsymbol\phi_{\boldsymbol\gamma}(\theta ; (Z^\prime, Y))\right].
\label{19}
\end{equation}
\label{t1}
\end{theorem}
\par Based on Theorem \ref{t1}, we reformulate problem (\ref{10}) and obtain the DRO-LS model as follows.
\begin{equation}
    \min_{\theta \in \Theta}E_{(Z^\prime,Y) \sim \hat{P_0}}\left[\boldsymbol\phi_{\boldsymbol\gamma}(\theta ; (Z^\prime, Y))\right],
    \label{100}
\end{equation}
where $\hat{P_0}$ is the empirical distribution of the data.
\subsection{Theoretical Analysis}\label{s:methods.3.2}\
To solve the penalty problem (\ref{100}), based on the definition of robust surrogate loss, we perform stochastic gradient descent on the robust surrogate loss \(\boldsymbol\phi_{\boldsymbol\gamma}\left(\theta ;\left(z^\prime, y\right)\right)\), and have:
\begin{equation}
\nabla_\theta {\boldsymbol\phi}_{\boldsymbol\gamma}\left(\theta ;\left(z^\prime, y\right)\right)=\nabla_\theta \boldsymbol\ell\left(\theta ;\left(z^{\prime\star}, y\right)\right),
\label{11}
\end{equation}
where \(z^{\prime\star} = \arg\max_{z \in \mathcal{Z}} \left\{\boldsymbol{\ell}\left(\theta ; (z, y_0)\right) - \gamma c_\theta\left((z, y_0), (z^\prime, y_0)\right)\right\}\) is a perturbation of the feature map \(z^\prime\) in the current parameters \(\theta\). In fact, we aim to obtain the \(z^{\prime\star}\) in the supremum stage. Supposing that \(\|z - z^\prime\|\) is sufficiently smooth. The following theorem illustrates that the relationship between \(z^\prime = f(\theta_r; x)\) and \(z^{\prime\star} = f(\theta_r; x^\star)\), where \(x\) denotes the existing training data and \(x^{\prime\star}\) is the optimal generated data. Specifically, we represent the non-identity Hessian matrix as $H$. 
\begin{theorem}[Relationship between Existing Data and Optimal Generated Data]
For \(z^\prime = f(\theta_r; x)\) and \(z^{\prime\star} = f(\theta_r; x^\star)\), we have
\begin{equation}
\begin{aligned}
& \left\|z^{\prime\star} - z^\prime\right\| = \left\|(H - \gamma I)^{-1} \nabla_{z^{\prime}}\boldsymbol\ell(\theta ;(z^{\prime}, y))\right\|,
\end{aligned}
\label{12}
\end{equation}
\label{t2}
\end{theorem}
where 
\begin{equation}
    \nabla_{z^{\prime}}\boldsymbol\ell(\theta ;(z^{\prime}, y)) = \left((1-\alpha) \theta_{f, i}^{\top}+\frac{\alpha}{(k-1)} \sum_{j \neq i}^k \theta_{f, j}^{\top}-\sum_{j=1}^k p_j(\theta ; z^{\prime}) \cdot \theta_{f, j}^{\top}\right)
\label{13}
\end{equation}
\par Theorem \ref{t2} reveals that the feature map of the optimal generated data \(z^{\prime\star}\) can be obtained by perturbing the feature map of the existing data \(z^\prime\). The mass transported from \(z^\prime\) to \(z^{\prime\star}\) in the semantic space is equal to the product of the deviation of \(\boldsymbol\ell(\theta ;(z^{\prime}, y))\) and the Hessian matrix. Theorem \ref{t2} further provides the detailed expression for \(\nabla_{z^\prime}\boldsymbol{\ell}(\theta ;(z', y))\), as shown in equation (\ref{13}), which consists of the feature map \(z^\prime\), the classification layer parameters $\theta_f$. Note that the hyperparameter in equation (\ref{5}) applied to labels now applies to \(\theta_f\). For the parameters \(\theta_{f,i}\) corresponding to the true category $i$, the weight is \(1-\alpha\) while the weights correspond to other positions are \(\frac{\alpha}{k-1}\). Instead of perturbations $L_p$ for the training samples, this term aims to penalize the distance between \((1-\alpha)\theta_{f, i}^\top + \frac{\alpha}{(k-1)}\sum_{j\neq i}^{k}\theta_{f, j}^\top\), with the weight characterized by the hyperparameter \(\alpha\), and \(\sum_{j=1}^{k} p_j(\theta ; z^\prime) \cdot \theta_{f, j}^\top\), which is the estimated mean of the $\theta_f$ across all categories with the feature map $z^\prime$. When $\alpha = 0$, the equality (\ref{13}) actually degenerates the results of the general DRO for cross entropy loss. As \(\alpha\) increases, the term gradually shifts focus towards the weights of the classification layer parameters corresponding to the remaining \(k-1\) categories, while somewhat reducing the emphasis on the weight of the parameters for the true label. Thus, the process of adjustment generating samples has been achieved by only adjusting the LS hyperparameter for $\theta_f$.
\par We now introduce the following \(L\)-Lipschitz continuity condition to ensure that the robust surrogate loss \(\boldsymbol{\phi}_{\boldsymbol{\gamma}}(\theta ; (z^\prime, y))\) can be effectively tractable (\citeauthor{oberman2019lipschitz}, \citeyear{oberman2019lipschitz}).
\begin{assumption}[\(L\)-Lipschitz Continuity, \citeauthor{sinha2018certifiable}, \citeyear{sinha2018certifiable}]
There exist \(L_0, L_1 > 0\) such that for all \(z, z^\prime\ \in \mathcal{Z}\) and \(\theta, \theta^\prime \in  \Theta\), we have
\begin{equation*}
\begin{aligned}
&\left\|\boldsymbol\ell\left(\theta ;\left(z, y\right)\right)-\boldsymbol\ell\left(\theta ;\left(z^{\prime}, y\right)\right)\right\| \leq L_{0}\left\|z-z^{\prime}\right\|_{2}, 
\left\|\nabla_{\theta} \boldsymbol{\ell}\left(\theta ;\left(z, y\right)\right)-\nabla_{\theta} \boldsymbol{\ell}\left(\theta ;\left(z^{\prime}, y\right)\right)\right\|_{2} \leq L_{1}\left\|z-z^{\prime}\right\|_{2}, \\
&\left\|\nabla_{\theta}\boldsymbol{\ell}(\theta^\prime;(z^{\prime\star},y)) - \nabla_{\theta}\boldsymbol{\ell}(\theta;(z^{\prime\star},y))\right\|_2 \leq L_\theta\|\theta^\prime-\theta\|_2.
\end{aligned}
\end{equation*}
\label{l4}
\end{assumption}
\par Let $f_{0}(z)=-\boldsymbol\ell_{1}\left(\theta ;\left(z, y\right)\right)+\frac{\gamma}{2}\left\|z-z^\prime\right\|_{2}^{2}$, $f_{1}(z)=-\boldsymbol\ell\left(\theta ;\left(z, y\right)\right)+\frac{\gamma}{2}\left\|z-z^\prime\right\|_{2}^{2}$. We present the bound for $z$ and $z^\prime$ in the following.
\begin{lemma}[\citeauthor{bonnans2013perturbation}, \citeyear{bonnans2013perturbation}]
If $f_{0}$, $f_{1}$ satisfies the second order growth condition, there exists a neighborhood $N$ of $z^{\prime}$ and a constant $\gamma>0$ for $f_{0}$ and $f_{1}$, such that
\begin{equation*}
f_{0}(z) \geq f_{0}(z^{\prime})+\gamma\left\|z-z^{\prime}\right\|_{2}^{2}.
\label{20}
\end{equation*}   
\label{l5}
\end{lemma}
\par Now, we obtain the bound on equality (\ref{12}) in Theorem \ref{t2}. 
\begin{theorem}[Bound of Difference between Existing Data and Optimal Generated Data]
    Let Assumption \ref{l4} hold. For \(z^\prime = f(\theta_r; x)\) and \(z^{\prime\star} = f(\theta_r; x^\star)\), we get
\begin{equation*}
\left\|z^{\prime\star}-z^{\prime}\right\|_{2} \leq \frac{3L_0}{\gamma}
\end{equation*}
\label{t6}   
\end{theorem}
\par Even though the variation of \(\alpha\) can further shift the worst-case distribution and modify the generated feature map \(z^{\prime\star}\), Theorem \ref{t6} demonstrates that there exists a bound to limit this shift.
\par Using Taylor theorem, we obtain
\begin{equation}
\begin{aligned}
\boldsymbol\ell\left(\theta ;\left(z, y\right)\right) = \boldsymbol\ell\left(\theta ;\left(z^\prime, y\right)\right) + \nabla_{z^\prime}\boldsymbol\ell\left(\theta ;\left(z^\prime, y\right)\right)^T(z-z^{\prime}) + O(\|z - z^\prime\|_2^2).   
\end{aligned}
\label{70}
\end{equation}
\par Since \(O(\|z - z^\prime\|_2^2)\) represents a higher-order term, we replace it with \(\frac{H}{2}\|z - z^\prime\|_2^2\), where \(H\) is a non-identity Hessian matrix. Based on equality (\ref{70}), the robust surrogate loss $\boldsymbol\phi_{\gamma}(\theta ;(z^\prime, y))$ can be reformulated as follows.
\begin{equation}
\begin{aligned}
\boldsymbol{\phi_{\gamma}}(\theta ;(z^\prime, y))=\sup _{z \in \mathcal{Z}}\{\boldsymbol\ell\left(\theta ;\left(z^\prime, y\right)\right) + \nabla_{z^\prime}\boldsymbol\ell\left(\theta ;\left(z^\prime, y\right)\right)^T(z-z^{\prime}) + \frac{H-\gamma I}{2}\|z - z^\prime\|_2^2\}.
\end{aligned}
\label{80}
\end{equation}
\par Based on Theorem \ref{t2}, the closed form of the surrogate robustness loss \(\boldsymbol{\phi}_{\boldsymbol{\gamma}}(\theta ;(z^\prime, y))\) can be formulated as follows:
\begin{equation}
    \boldsymbol\phi_{\boldsymbol\gamma}(\theta ;(z^\prime, y)) = \boldsymbol\ell\left(\theta ;\left(z^\prime, y\right)\right) + \frac{(\gamma I - H)^{-1}}{2} \|\nabla_{z^\prime}\boldsymbol\ell\left(\theta ;\left(z^\prime, y\right)\right)\|_2^2
    \label{14}
\end{equation}
\par \(\nabla_{z^\prime}\boldsymbol\ell\left(\theta ;\left(z^\prime, y\right)\right)\) in equation (\ref{14}) can be regarded as an \(L_2\) regularization term, which integrates the regularization effect of LS. Since $\|z - z^\prime\|$ is smooth enough, the Hessian matrix $H$ is sparse. Following \citeauthor{volpi2018generalizing} (\citeyear{volpi2018generalizing}), we further use $\frac{1}{\gamma}$ to approximately replace $\frac{(\gamma I - H)^{-1}}{2}$ in equality (\ref{14}) and obtain:
\begin{equation}
    \boldsymbol\phi_{\boldsymbol\gamma}(\theta ;(z^\prime, y)) = \boldsymbol\ell\left(\theta ;\left(z^\prime, y\right)\right) + \frac{1}{\gamma}\|\nabla_{z^\prime}\boldsymbol\ell\left(\theta ;\left(z^\prime, y\right)\right)\|_2^2
    \label{15}
\end{equation}
\par Observe that the parameter $\gamma$ in equality (\ref{15}) also participates in the data shift process. Therefore, perturbations on $\gamma$ also influences the existing data shift to unseen domains. The following bound describes such perturbations in a more detailed way. 
\begin{theorem}[Bound of Surrogate Robustness Loss]
We define $L(\theta) = \sum_{j=1}^{k}\|\theta_{f,j}\|_2\max_{1 \leq j^{'} \leq k}\|\theta_{f, j^{'}}\\ \|_2$. If $\gamma>L(\theta)$, we have the following bound:
\begin{equation}
\begin{aligned}
&\boldsymbol\phi_{\boldsymbol\gamma}(\theta ; (z^\prime, y))\leq \boldsymbol\ell(\theta ;(z^\prime, y))  + \frac{1}{\gamma-L(\theta)}\|(1-\alpha)\theta_{f, i}^{\top}+\frac{\alpha}{(k-1)}\sum_{j\neq i}^{k}\theta_{f, j}^{\top}-\sum_{j=1}^{k} p_{j}(\theta ; z) \cdot \theta_{f, j}^{\top}\|_{2}^{2},
\end{aligned}
\label{16}
\end{equation}
\begin{equation}
\begin{aligned}
&\boldsymbol\phi_{\boldsymbol\gamma}(\theta ; (z^\prime, y)) \geq
\boldsymbol\ell(\theta ;(z^\prime, y)) + \frac{1}{\gamma+L(\theta)}\|(1-\alpha)\theta_{f, i}^{\top} + \frac{\alpha}{(k-1)}\sum_{j\neq i}^{k}\theta_{f, j}^{\top}-\sum_{j=1}^{k} p_{j}(\theta ; z^\prime) \cdot \theta_{f, j}^{\top}\|_{2}^{2}.   
\end{aligned}
\label{17}
\end{equation}
\label{t7}
\end{theorem}
\par Theorem \ref{t7} illustrates that the bound on $\boldsymbol\phi_{\boldsymbol\gamma}(\theta ; (z^{\prime}, y))$ is equivalent to the sum of $\boldsymbol\ell(\theta ;(z^{\prime}, y))$ and \(\nabla_{z^\prime}\boldsymbol\ell\left(\theta ;\left(z^\prime, y\right)\right)\), with distinct perturbations $L(\theta)$ for $\gamma$. Note that the terms of the bound are the same as the regularization term in equality (\ref{16}). Therefore, the bound can be regarded as an indicator that measures the quality of generating samples consistent with equation (\ref{15}). If the performance of DNNs trained using generated samples associated with the bound, drops dramatically, it indicates that consistently generating samples as described in equality (\ref{15}) is considerably difficult. Otherwise, generating the same samples is relatively easy. 
\subsection{Proposed Algorithm}\label{sec:3.3}\
In this section, based on the above theoretical analysis, we construct the two-stage algorithm GI-LS for problem (\ref{100}), as detailed in Algorithm (\ref{alg:Framwork}). Specifically, GI-LS includes two stages: $(1)$ the inner maximization stage, which utilizes the regularization effect of LS to shift data to unseen domains and generate new samples, and $(2)$ the outer minimization stage, which trains DNNs using the generated samples.
\begin{algorithm}[htb]  
  \caption{GI-LS} 
  \label{alg:Framwork}  
  \begin{algorithmic} 
  \setlength{\baselineskip}{0.7\baselineskip} 
    \Require
      small-scale dataset $\left\{X_{i}, Y_{i}\right\}_{i=1}^{N}$ and initialized weights $\theta$  
    \Ensure 
      trained weights $\theta$
    \For{$k = 1$, $\ldots$, $K$}
    \For{$i = 1$, $\ldots$, $N$}
    \State selecting sample $\left(X_{i}, Y_{i}\right)$ from dataset $\left\{X_{i}, Y_{i}\right\}_{i=1}^{N}$
    \State $X_i^t \leftarrow X_i$
    \For{t = 1, $\ldots$, T}
      \State $X^{t+1}_{i} \leftarrow X^{t}_{i}+\eta\nabla_{x}\{\boldsymbol\ell\left(\theta ;\left(X^{t}_{i}, Y_{i}\right)\right) - \gamma c_{\theta}\left(\left(X^{t}_{i}, Y_{i}\right),\left(X_{i}, Y_{i}\right)\right)\}$
    \EndFor
    \EndFor
    \State $\theta^{\prime k} \leftarrow \theta^{k} -\beta \nabla_{\theta}\boldsymbol\ell(\theta ;(X_{i}^T, Y_{i}))$
    \State $\theta^{k+1} \leftarrow \theta^{\prime k} -\beta \nabla_{\theta}\boldsymbol\ell(\theta ;(X_{i}, Y_{i}))$
    \EndFor
  \end{algorithmic}  
\end{algorithm}
\par As seen from the above review, the inner supremum of problem (\ref{100}) approximately equals to equality (\ref{15}). However, the term \(\nabla_{z^{\prime}}\boldsymbol\ell(\theta ;(z^{\prime}, y))\) involves complex and dense matrix, making direct solution difficult. Following \cite{gao2017distributional}, we consider the gradient ascent algorithm to approximate the quantity of the term. In fact, equality (\ref{12}) in Theorem \ref{t2} provides the theoretical motivation for applying gradient iteration scheme to address the inner supremum of the problem (\ref{100}). Specifically, in the inner maximization stage of GI-LS, we employ stochastic gradient ascent to approximate the supremum of the robust surrogate loss \(\boldsymbol\phi_{\boldsymbol\gamma}(\theta ;(z^\prime, y))\) over multiple iterations. The intensity of the worst-case distribution is controlled by the parameter $\alpha$ in LS. After iterations, the regularization term in equation (\ref{13}) introduces perturbations to the existing data, generating new data.
\par For the outer minimization in problem (\ref{100}), we consider the stochastic gradient descent method to solve it. Specifically, in the outer minimization stage of GI-LS, both the data generated in the supremum stage and the original data serve as the training samples. Additionally, we continue to use LS as the loss function for training DNNs in this stage.
\par The following theorem guarantees that GI-LS still can make the parameters of the classification layer converge to a stationary point. Let $g\left(\theta, z; z^{\prime}\right) = \sup_{z \in \mathcal{Z}}\boldsymbol \ell\left(\theta ;\left(z, y\right)\right)-\frac{\gamma}{2}\left\|z-z^{\prime}\right\|_2^2$. 
Noting $F(\theta)=\operatorname{E}_{{P}_{0}}\left[\boldsymbol\phi_{\boldsymbol\gamma}(\theta ; (z^\prime, y))\right]$. Let $\Delta_{F} \geq F\left(\theta^{1}\right)-\inf _{\theta} F(\theta)$. The convergence of GI-LS is illustrated in the following.
\begin{theorem}[Algorithm Convergence] 
Let Assumption \ref{l4} hold. We assume \small{${E}\left[\left\|\nabla F(\theta)\-\nabla_{\theta} \boldsymbol{\phi_{\gamma}}(\theta, (z,y))\right\|_{2}^{2}\right] \leq\sigma^{2}$} and take a constant step size $\beta = \sqrt{\frac{2\Delta_{F}}{L_{\theta}K\sigma^2}}$. For $K \geq \sqrt{\frac{L_\theta\Delta_{F}}{2\sigma^2}}$, GI-LS satisfies\\
\begin{equation}
\begin{aligned}
\frac{1}{K}\sum_{k=1}^{K} {E}\left[\left\|\nabla F\left(\theta^{k}\right)\right\|_{2}^{2}\right] - \frac{3L_0L_1^2}{\gamma}\leq \sigma\sqrt{\frac{18L_\theta\Delta_{F}}{K}}  
\end{aligned}
\label{18}
\end{equation}
\label{t8}
\end{theorem}
\par The perturbation on the feature map $z^\prime$ is independent of the iteration count $K$, where the gradient ascent method applied in the supremum stage does not affect the convergence rate of problem (\ref{100}) even if we integrate LS. In fact, Theorem \ref{t8} illustrates that problem (\ref{100}) is equivalent to that of standard smooth non-convex optimization (\citeauthor{ghadimi2013stochastic}, \citeyear{ghadimi2013stochastic}; \citeauthor{sinha2018certifiable}, \citeyear{sinha2018certifiable}). 
\section{Case Study}\label{sec:4}\
In this section, we apply GI-LS to a real-world case study on the classification task for small-scale anomaly images with DNNs, which includes two magnetic tile surface defect datasets (\citeauthor{huang2020surface}, \citeyear{huang2020surface}) and two types of textures (wood and carpet) with anomalies from the MVTec AD dataset (\citeauthor{bergmann2019mvtec}, \citeyear{bergmann2019mvtec}), to validate the effectiveness of the proposed GI-LS. In Section \ref{sec:4.1}, we provide information on the datasets for magnetic tile surface defects and the datasets for wood and carpet anomalies. In Section \ref{sec:4.2}, we evaluate the classification performance of DNNs with BO for GI-LS and compare it with the general LS. Furthermore, we use four datasets as instances to analyze the hyperparameter selection for real-world small-scale anomaly image classification tasks. In Section \ref{app9}, we compare GI-LS with the common existing data augmentation methods. Moreover, we validate the performance of GI-LS after the existing samples are attacked by common adversarial data augmentation methods. In Section \ref{app10}, we also discuss the results of perturbation on $\gamma$ for generating samples.
\subsection{Small-scale Anomaly Dataset}\label{sec:4.1}\
We use magnetic tile surface defect (\citeauthor{huang2020surface}, \citeyear{huang2020surface}), following the structure of CUB-200-2011 (\citeauthor{wah2011caltech}, \citeyear{wah2011caltech}), to establish two magnetic tile surface datasets, as shown in Table \ref{Table2}. \textit{MT-Defect} includes five types of defects. In contrast, \textit{MT} introduces images without surface defects, which are far more numerous than the defect categories. In general, the number of products with surface defects is usually less than the number of qualified products in a real-world industrial line, and the number of each type of surface defect product is rare. The distribution of \textit{MT} is consistent with the real-world situation to some extent.
\begin{table*}[ht]
\renewcommand\arraystretch{0.7}
\caption{Introduction of magnetic tile surface defect dataset}
\label{Table2}
\centering
\setlength{\tabcolsep}{4.0mm}{\begin{tabular}{cccccccccccccccc}
\toprule
Dataset &Category & Free & Uneven & Blowhole & Break & Crack & Fray \\
\hline
\bfseries \textit{MT-Defect} &Training &- &54 &67 &46 &28 &13 \\
\bfseries &Testing (unseen)  &- &49 &48 &39 &26 &16 \\
\bfseries \textit{MT} &Training  &460 &41 &57 &52 &27 &13 \\
\bfseries &Testing (unseen)  &492 &62 &58 &33 &27 &16 \\
\bottomrule
\end{tabular}}
\end{table*}
\par
\begin{figure*}[ht]
\centering
\includegraphics[width=6.5in]{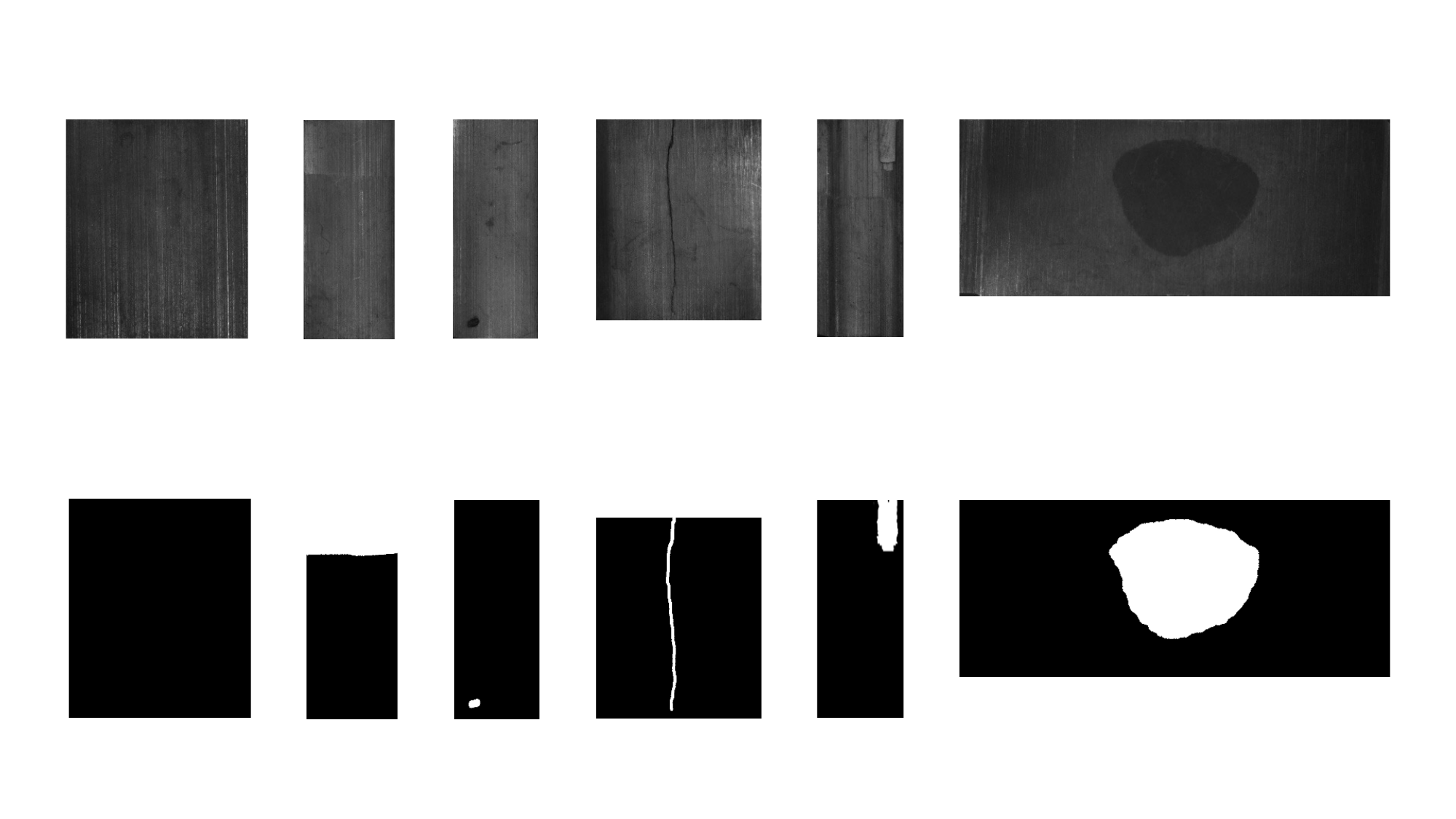}
\caption{Some surface defects of magnetic tiles.}
\label{Figure3}
\end{figure*}

 We provide some samples for \textit{MT-Defect} and \textit{MT} in Figure \ref{Figure3}. The first row from left to right corresponds to `Free', `Uneven', `Blowhole', `Break', `Crack', and `Fray', respectively. The second row is the position of the surface defect.
\par Moreover, we also select two anomalies, wood and carpet, from the MVTec AD dataset (\citeauthor{bergmann2019mvtec}, \citeyear{bergmann2019mvtec}) to establish two surface anomaly datasets, \textit{Wood} and \textit{Carpet}, as illustrated in Table \ref{Table3}. Both datasets contain a series of anomalies (see Table \ref{Table3}). The distributions of the datasets are similar to \textit{MT}, where the samples without anomalies are more numerous than the anomaly samples.
\begin{table*}[htbp]
\renewcommand\arraystretch{0.7}
\caption{Introduction of surface anomaly dataset}
\label{Table3}
\centering
\setlength{\tabcolsep}{0.5mm}{\begin{tabular}{cccccccccccccccc}
\toprule
Dataset &Category & Free  &Color &Combined &Hole &Liquid &Scratch &Thread &Cut &Mental-contamination\\
\hline
\bfseries \textit{Wood} &Training  &110  &4 &4 &3 &4 &12 &- &- &- \\
\bfseries &Testing (unseen)        &137  &4 &7 &7 &6 &9  &- &- &- \\
\bfseries \textit{Carpet} &Training  &167 &12 &- &12 &- &- &14 &10 &13 \\
\bfseries &Testing (unseen)          &113 &7  &- &5  &- &- &5  &7 &4 \\
\bottomrule
\end{tabular}}
\end{table*}
\par
\begin{figure*}[ht]
\centering
\hspace*{-0.5cm} 
\includegraphics[width=6.5in]{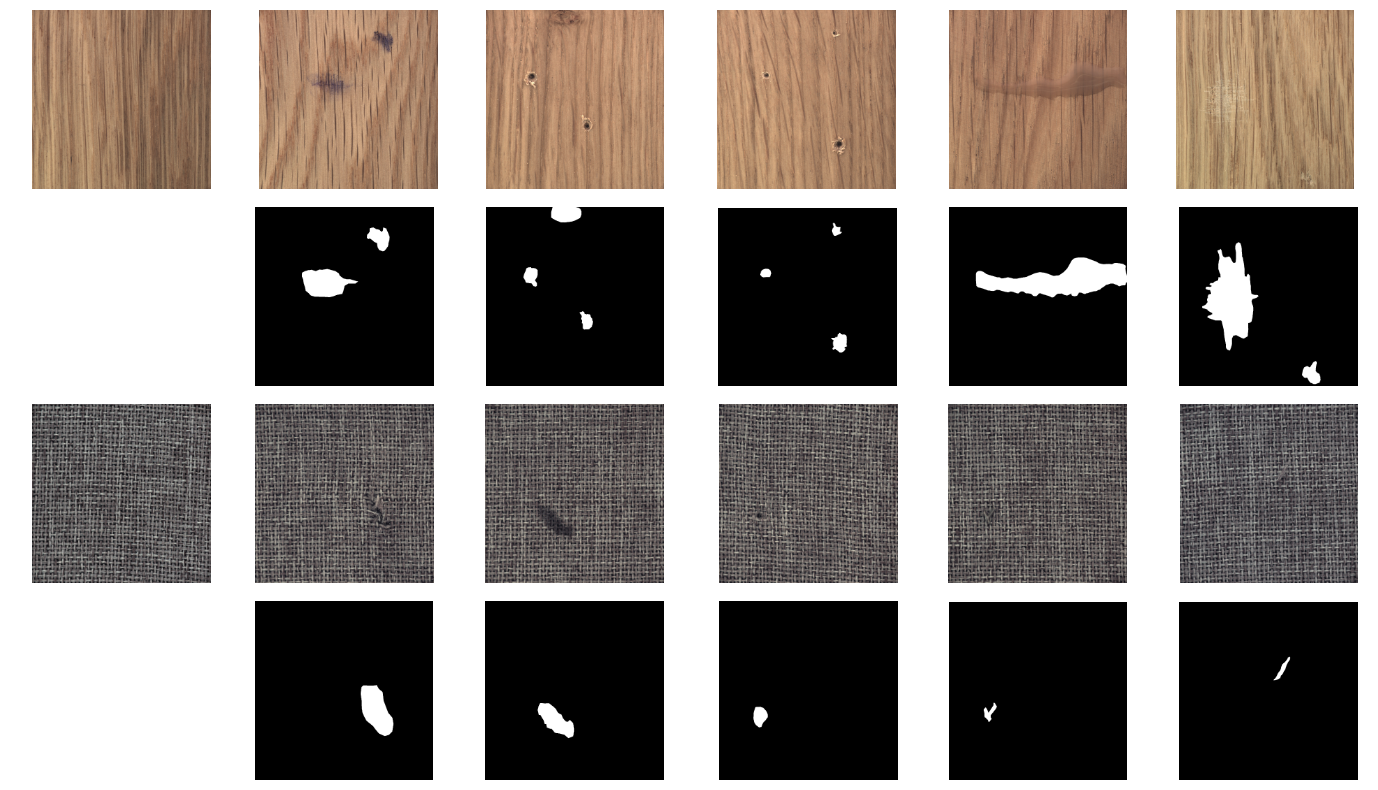}
\caption{Some surface anomalies of Wood and Carpet.}
\label{Figure4}
\end{figure*}

 Figure \ref{Figure4} show some instances for the samples in \textit{Wood} and \textit{Carpet}. The first row represents Wood. From left to right correspond to `Free', `Color', `Combined', `Hole', `Liquid', and `Scratch', respectively. The second row indicates the position of these surface anomalies. The third row shows some anomalies of Carpet. From left to right correspond to `Free', `Color', `Hole', `Cut', `Metal-contamination', and `Thread', respectively. The fourth row indicates the position of the surface anomalies.
\subsection{Bayesian Optimization}\label{sec:4.2}\
As seen from the above review, the proposed GI-LS includes a series of hyperparameters. When applying stochastic gradient ascent to generate new data for the inner maximization stage of GI-LS, in addition to the \(\alpha\) of LS, the number of perturbation iterations \(T\) and the step size of gradient ascent \(\eta\) are also crucial to the data generation process. Furthermore, the iterations of the outer minimization stage of GI-LS determine the efficiency. Thus, selecting appropriate ranges for these parameters is essential for the performance of GI-LS.
\par Bayesian Optimization (BO) has been widely used when the process of training DNNs involves multiple hyperparameters, especially in the context of manufacturing (\citeauthor{albahar2021robust}, \citeyear{albahar2021robust}). It is known for its efficacy in globally optimizing expensive black-box functions (\citeauthor{semelhago2021rapid}, \citeyear{semelhago2021rapid}; \citeauthor{xu2023constrained}, \citeyear{xu2023constrained}). Therefore, we employ BO to explore the optimal combination of the hyperparameters. Specifically, we use Adaptive Experimentation (Ax) platform of Meta to execute BO for GI-LS (\citeauthor{baird2022high},\citeyear{baird2022high}). To address the covariance of uncertainty, the Matern $5/2$ kernel is employed. We mainly focus on the parameters of the classification layer and treat the other hidden layers as the feature extraction layer. In our settings, we adopt ResNet architectures—\textit{ResNet18}, \textit{ResNet34}, and \textit{ResNet50}—as the feature extraction layers, which have been pre-trained on the ImageNet dataset (\citeauthor{he2016deep}, \citeyear{he2016deep}). 
\par In our experiment, the value of \(\gamma\) is set to \(10^{-3}\). We execute 20 iterations of BO for GI-LS. The perturbation factor \(\alpha\) of LS is bounded between 0 and 0.5, the number of iterations \(T\) ranges from 1 to 15, the step size \(\eta\) ranges from 1.5 to 3.0, and the number of training rounds ranges from 21 to 60. In each iteration, the learning rate \(\beta\) is set to \(1 \times 10^{-2}\) for the classification layer and \(1 \times 10^{-3}\) for the parameters of the feature extraction layer, reducing by a factor of 0.3 every 20 epochs. All experiments are conducted within the PyTorch framework on Nvidia GeForce RTX 3060 and 3090 GPUs.
\subsubsection{Classification Results}\label{sec:4.2.1}\
\vspace{0.2cm}
\begin{figure}[htbp]
\centering
\hspace*{-1.0cm} 
\begin{minipage}[t]{0.48\textwidth}
\centering
\includegraphics[width=8.5cm]{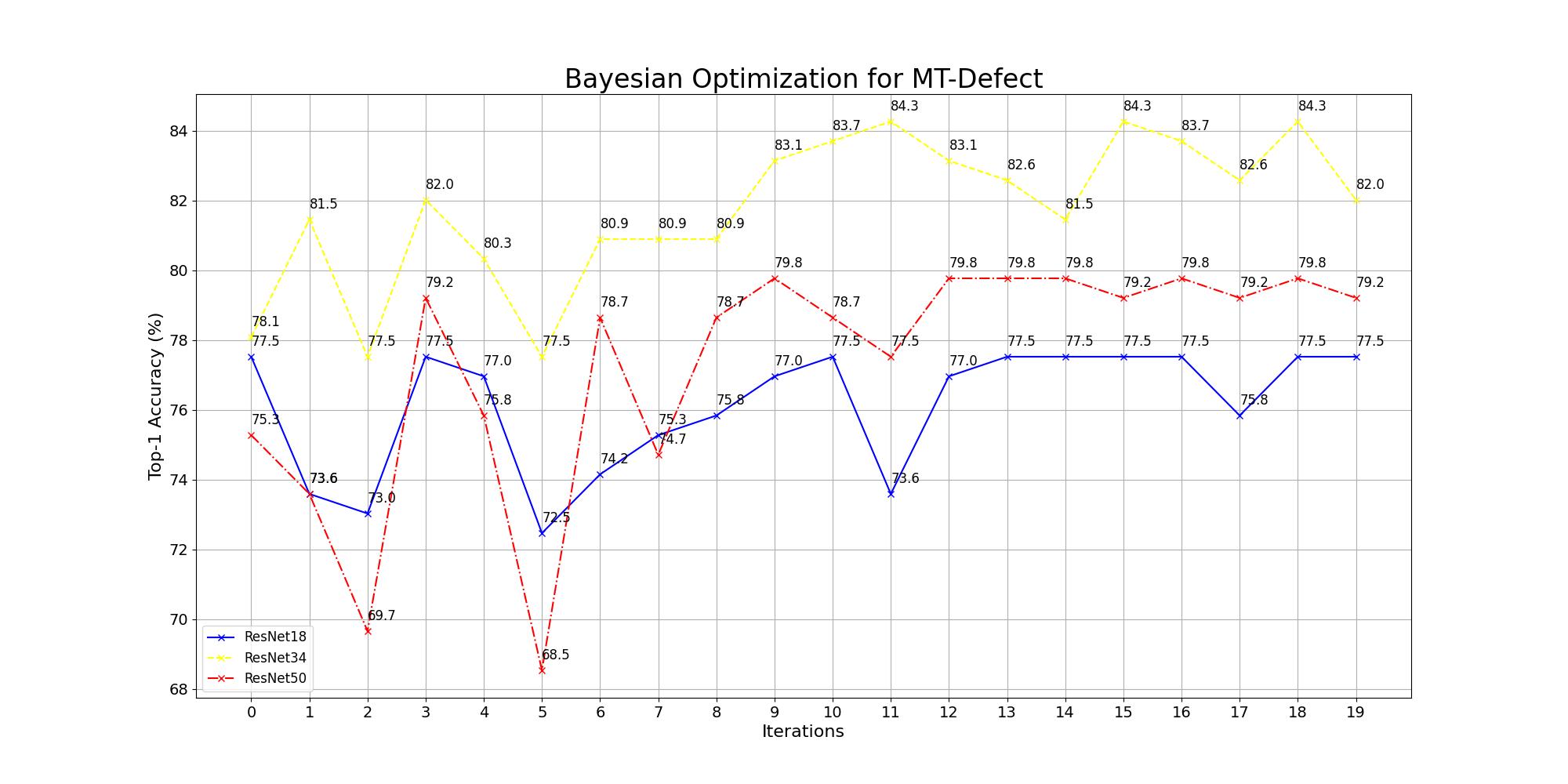}
\end{minipage}
\hspace*{0.5cm} 
\begin{minipage}[t]{0.48\textwidth}
\centering
\includegraphics[width=8.5cm]{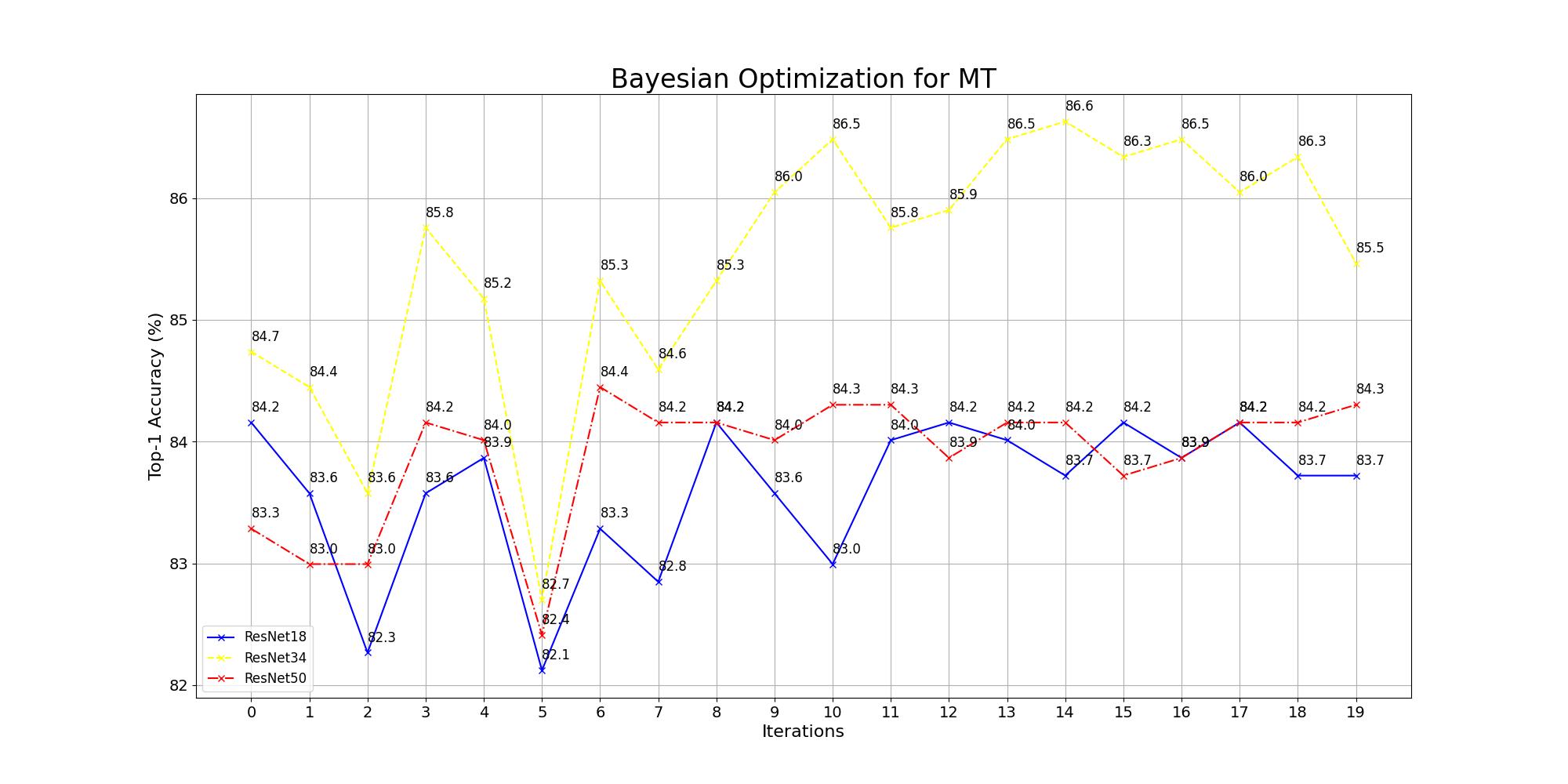}
\end{minipage}
\notag
\end{figure}
\begin{figure}[htbp]
\centering
\hspace*{-1.0cm} 
\begin{minipage}[t]{0.48\textwidth}
\centering
\includegraphics[width=8.5cm]{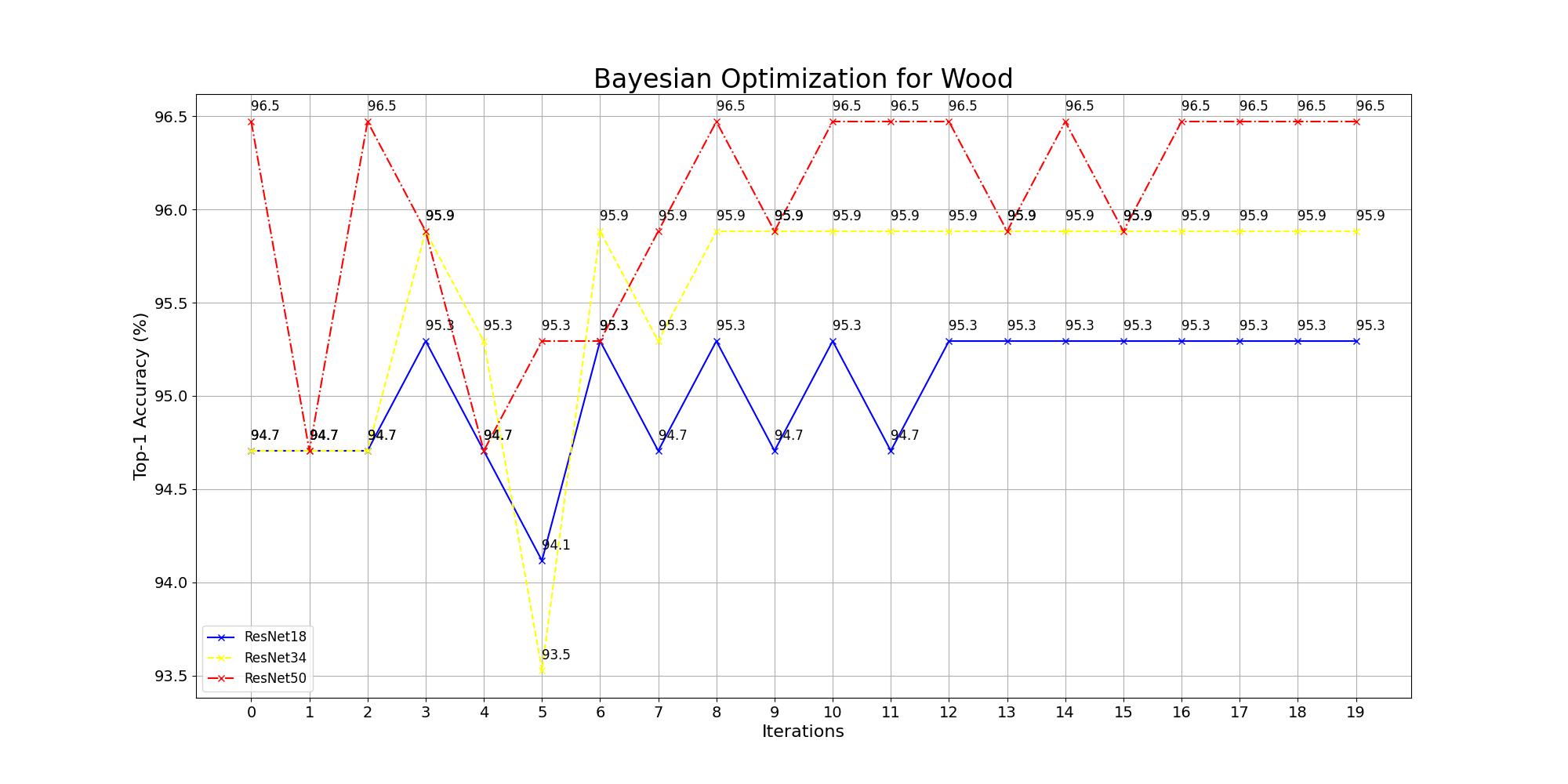}
\end{minipage}
\hspace*{0.5cm} 
\begin{minipage}[t]{0.48\textwidth}
\centering
\includegraphics[width=8.5cm]{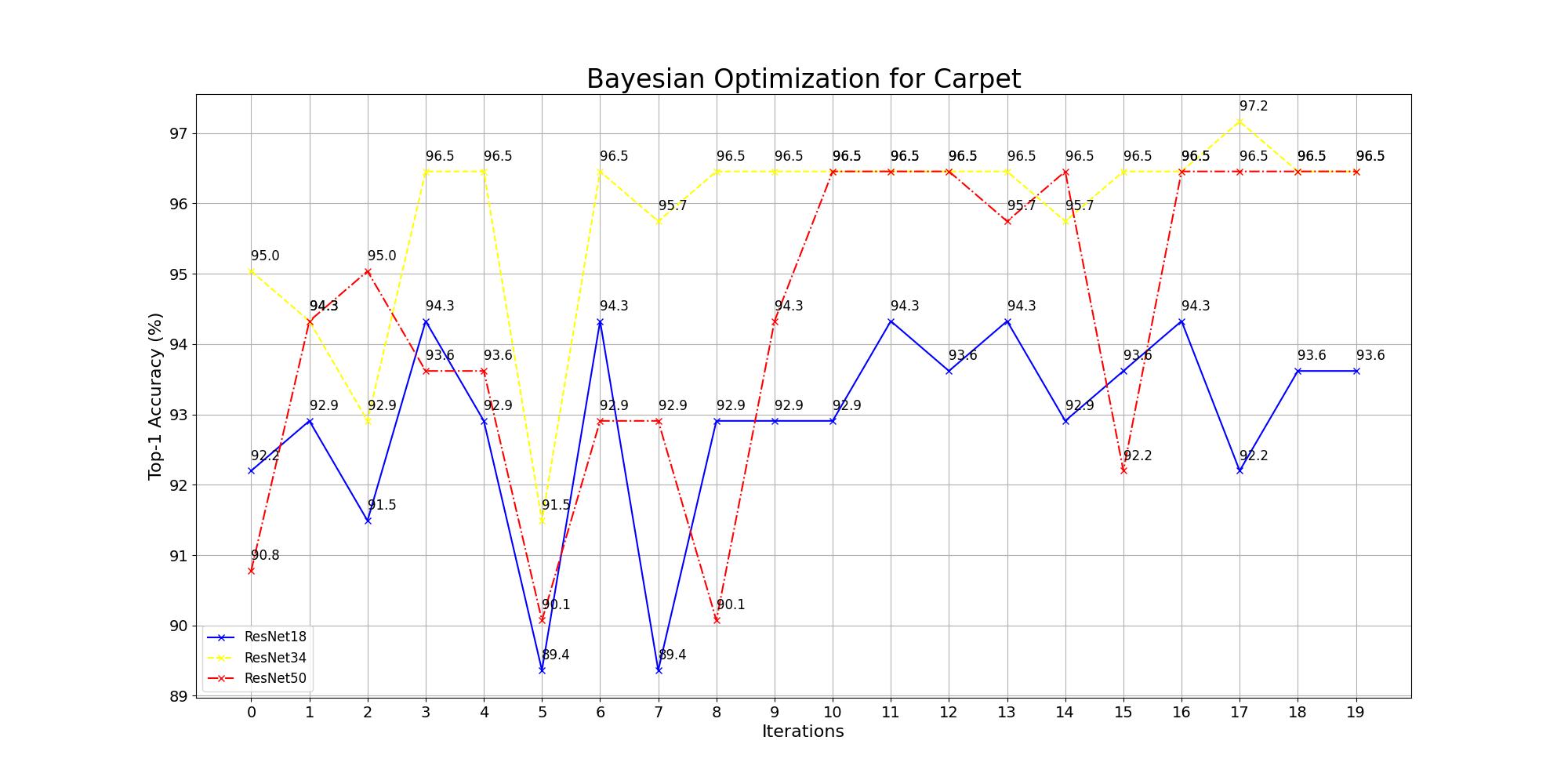}
\end{minipage}
\caption{Bayesian optimization for \textit{MT-Defect}, \textit{MT}, \textit{Wood} and \textit{Carpet}. The blue, yellow and red lines represent the results of \textit{ResNet18}, \textit{ResNet34}, and \textit{ResNet50}, respectively.}
\label{Figure1}
\end{figure}
\par Figure \ref{Figure1} presents the top-1 accuracy of DNNs with the execution of BO iterations across four datasets. Each point represents the top-1 accuracy of one iteration. \textit{ResNet34} achieves competitive results for all datasets, followed by \textit{ResNet50}, with \textit{ResNet18} coming last, demonstrating that selecting an appropriately-sized feature extractor is crucial when applying GI-LS to small-scale datasets. Compared with \textit{MT-Defect} and \textit{MT}, the classification performance of all DNNs improves on \textit{MT}, illustrating that increasing the number of perturbed `Free' samples can enhance the generalization of DNNs. Moreover, the trends observed for \textit{ResNet34} on \textit{MT-Defect} and \textit{MT} are consistent, as are those for \textit{ResNet18} and \textit{ResNet50}. For instance, the performance of all DNNs declines dramatically during the $6^{th}$ iteration in both \textit{MT-Defect} and \textit{MT}. This indicates that the search space of the hyperparameters exhibits some degree of similarity for all three types of DNNs. Additionally, the similar trends observed between \textit{MT-Defect} and \textit{MT} clarify that the increase in `Free' samples does not significantly affect the search space of the hyperparameters. For \textit{Wood} and \textit{Carpet}, GI-LS achieves superior performance for all three DNNs, with \textit{ResNet34} also achieving competitive results, highlighting the significance of selecting the appropriate feature-extractor size again. The structures of \textit{Wood} and \textit{Carpet} are the same as \textit{MT}, where the superior results further reflect the benefits of adding perturbed `Free' samples for the feature extraction process. Compared with the defects in \textit{MT}, the quantity of each type of anomaly in \textit{Wood} and \textit{Carpet} is small. The stable performance of DNNs in both \textit{Wood} and \textit{Carpet} demonstrates that the generalization of DNNs is excellent when applying GI-LS. Even if the performance of DNNs is good enough for \textit{Wood} and \textit{Carpet}, the slight fluctuation of DNNs performance reflects that adjusting \(\alpha\) in LS for generated samples can still impact model's generalization for some samples. In addition, the superior performance demonstrates that the search space for the hyperparameters is further flattened for \textit{Wood} and \textit{Carpet}. Therefore, we can focus on adjusting down the values of some hyperparameters that influence training efficiency (e.g., the number of perturbation iterations \(T\), and epochs).
\par In summary, we are supposed to select appropriate feature extractor based on the scale of datasets when applying GI-LS. The increasing of `Free' samples to dataset can also improve DNNs performance. If DNNs performance fluctuates significantly during BO iterations, as demonstrated with \textit{ResNet34} on \textit{MT-Defect} and \textit{MT}, we need to explore the selection space of the hyperparameters. Otherwise, the hyperparameters selections are flexible. In this situation, we can improve the training efficiency and maintain superior performance, as illustrated in \textit{Wood} and \textit{Carpet}.
\par We record the hyperparameter combinations when DNNs achieve optimal performance during BO iterations on \textit{MT-Defect}, \textit{MT}, \textit{Wood}, and \textit{Carpet}, as provided in Table \ref{Table101}.
\begin{table*}[htbp]
\renewcommand\arraystretch{0.7}
\caption{The optimal combination of hyperparameters after BO execution}
\label{Table101}
\centering
\setlength{\tabcolsep}{2.0mm}{\begin{tabular}{cccccccccccccccc}
\toprule
Dataset &DNNs &$\alpha$ &$T$ &$\eta$ &epoch &Accuracy($\%$) \\
\hline
\bfseries \textit{MT-Defect} &\textit{ResNet18}  &0.5 &15 &1.8830130484958096 &36 &77.5 \\
                             &\textit{ResNet34}  &0.40432489252683534 &10 &2.5677535344621507 &60 &84.3 \\
                             &\textit{ResNet50}  &0.2445120479817997  &7  &2.14415756835824   &58 &79.8 \\
\hline
\bfseries \textit{MT}        &\textit{ResNet18}  &0.40940070111766175 &15   &1.5  &42 &84.2 \\
                             &\textit{ResNet34}  &0.2959914206398052  &12   &2.495296771792481 &60 &86.6 \\
                             &\textit{ResNet50}  &0.13999486481770873 &6    &1.693113137036562 &48 &84.4 \\
                             \hline
\bfseries \textit{Wood}      &\textit{ResNet18}  &0.3659006094493754  &9 &2.164974745041862  &53   &95.3 \\
                             &\textit{ResNet34}  &0.17858423418416414 &6 &1.6735971654037183 &50   &95.9 \\
                             &\textit{ResNet50}  &0.37972310064635106 &12 &1.700566688241675 &43   &96.5 \\
                             \hline
\bfseries \textit{Carpet}    &\textit{ResNet18}  &0.06502356771215663 &7 &1.5648908500743228 &49   &94.3 \\
                             &\textit{ResNet34}  &0.3194096034244731  &8 &1.5 &60 &97.2 \\
                             &\textit{ResNet50}  &0.0                 &14 &2.822934206747137 &48   &96.5 \\
\bottomrule
\end{tabular}}
\end{table*}
\par Moreover, we select the optimal \(\alpha\) and epoch from Table \ref{Table101} to train DNNs with LS on \textit{MT-Defect}, \textit{MT}, \textit{Wood}, and \textit{Carpet}. The results are shown in Table \ref{Table100}.
\begin{table*}[htbp]
\renewcommand\arraystretch{0.7}
\caption{The performance of DNNs with LS ($\%$)}
\label{Table100}
\centering
\setlength{\tabcolsep}{9.0mm}{\begin{tabular}{ccccccc}
\toprule
&              &\textit{MT-Defect} &\textit{MT} &\textit{Wood} &\textit{Carpet} \\
\hline
&\textit{ResNet18}  &62.4 &75.0   &91.2 &95.7 \\
&\textit{ResNet34}  &64.6 &76.0   &91.2 &90.8 \\
&\textit{ResNet50}  &74.7 &76.3   &91.2 &95.0 \\
\bottomrule
\end{tabular}}
\end{table*}
\par Compared with LS, the performance of GI-LS shows significant improvement for the two magnetic tile surface defect datasets, while the performance of GI-LS on \textit{Wood} and \textit{Carpet} shows a certain degree of improvement. For \textit{MT-Defect} and \textit{MT}, the results demonstrate that the perturbations to existing data by LS integrated with DRO can generate data containing more effective information based on the existing data, improving the model's generalization. For \textit{Wood} and \textit{Carpet}, even though LS has already achieved great performance, the generated samples can still benefit the model's generalization for some samples. The reason that GI-LS performs better than the general LS can be summarized as follows: LS only considers multiple labels for the same data, while GI-LS considers multiple labels for multiple data. In GI-LS, DRO incorporating Wasserstein distance ensures that the generated data are distributed around the original data, while LS further helps to adjust the distribution of the generated data flexibly, fitting the training process. 
\subsubsection{Analysis of Hyperparameter Combination in GI-LS}\label{app8}\
\par As seen from the above review, the performance of DNNs on \textit{Wood} and \textit{Carpet} is relatively stable, indicating that the search space of the hyperparameters is flattened. In contrast, the performance of DNNs on \textit{MT-Defect} and \textit{MT} fluctuates to some extent. In this section, we analyze how the combination of hyperparameters in GI-LS clusters in the optimal region during the BO iteration for the four datasets.
\begin{figure*}[htbp]
\centering
\includegraphics[width=6.5in]{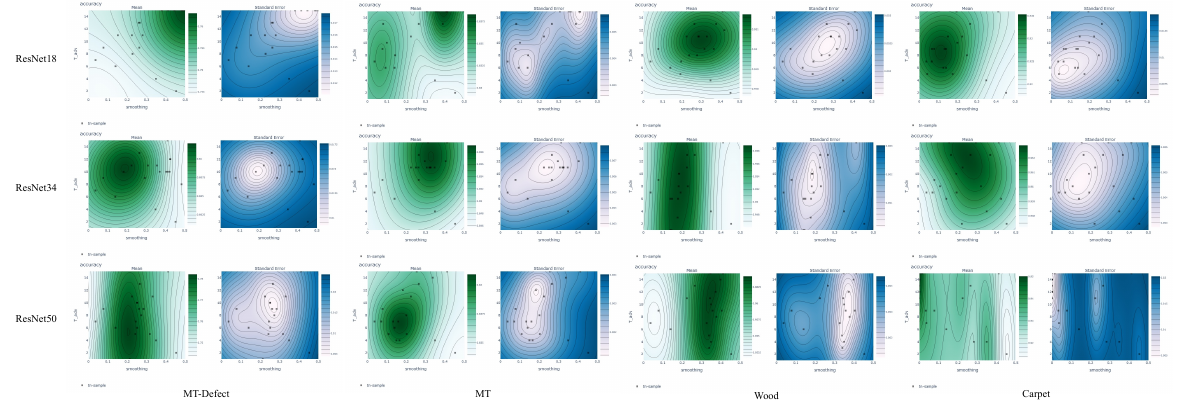}
\caption{The relationship between $\alpha$ and $T$ as BO iterates. From up to down corresponds to \textit{ResNet18}, \textit{ResNet34}, and \textit{ResNet50}. From left to right corresponds to \textit{MT-Defect}, \textit{MT}, \textit{Wood}, and \textit{Carpet}, respectively. The optimization results of each iteration are represented by black points. For each dataset, the left columns in green represent the top-1 accuracy for each BO iteration. The more concentrated the points, the higher the top-1 accuracy. The right columns illustrate the standard error for BO iteration. The more concentrated the points, the lower the standard error.}
\label{Figure5}
\end{figure*}
\par The relationship between $\alpha$ and $T$ is shown in Figure \ref{Figure5}. Note that \(\alpha\) does not cluster around the origin of the coordinate axis, indicating the effectiveness of the further shift for the worst-case distribution by LS. On \textit{MT-Defect} and \textit{MT}, a large combination of $\alpha$ and $T$ benefits the performance of \textit{ResNet18}, demonstrating the noteworthy role of $\alpha$ in shifting the worst-case distribution to achieve better performance. The results of \textit{ResNet34} and \textit{ResNet50} illustrate that decreasing \(T\) tends to achieve better performance with the increase in DNNs size. At the same time, larger $\alpha$ benefits \textit{ResNet34} while smaller $\alpha$ benefits \textit{ResNet50}. On \textit{Wood} and \textit{Carpet}, a smaller $T$ can maintain good performance of DNNs while improve the training efficiency. At the same time, $\alpha$ is sensitive to the \textit{Wood} and \textit{Carpet} with difference size DNNs. Since \textit{ResNet34} achieves competitive performance on all four datasets, we focus on its hyperparameter combination and find that a moderately sized combination of \(\alpha\) and \(T\) tends to achieve better performance and improve training efficiency.
\begin{figure*}[htbp]
\centering
\includegraphics[width=6.5in]{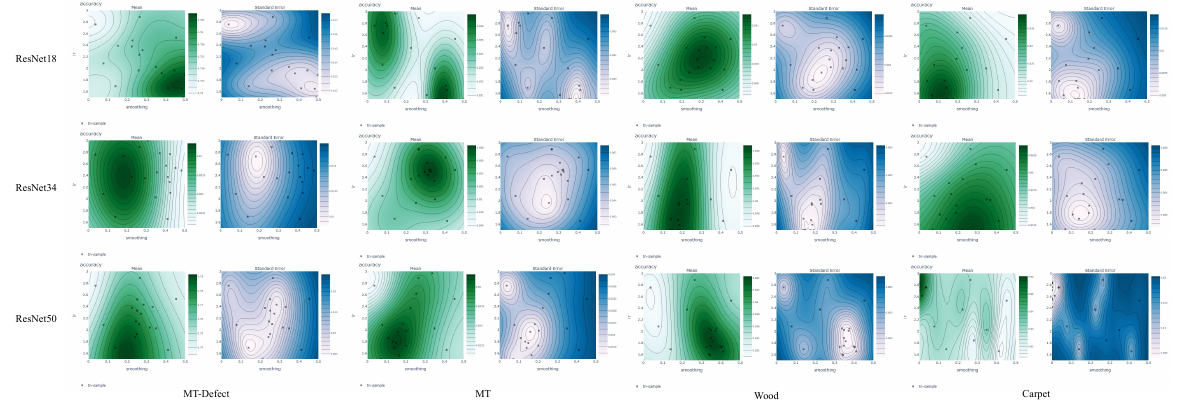}
\caption{The relationship between $\alpha$ and $\eta$ as BO iterates. The settings are the same as those in Figure \ref{Figure5}.}
\label{Figure6}
\end{figure*}
\par
\begin{figure*}[htbp]
\centering
\includegraphics[width=6.5in]{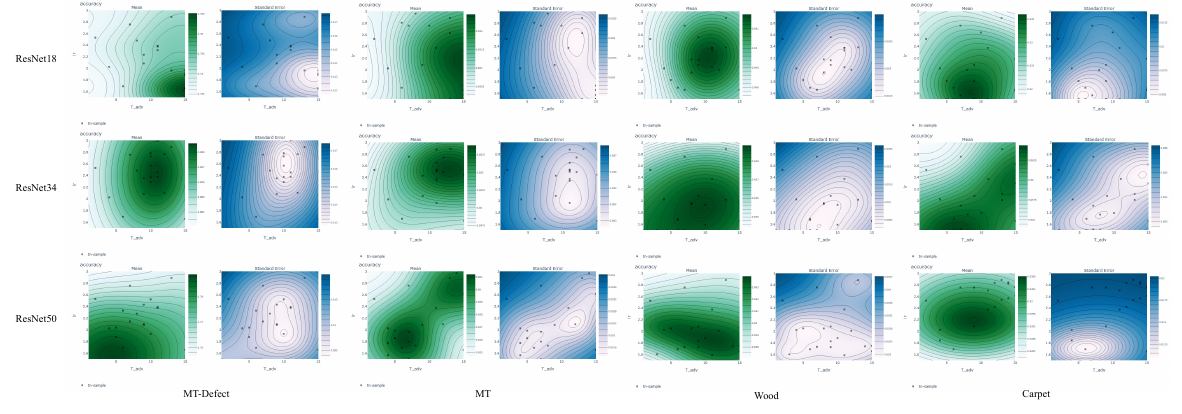}
\caption{The relationship between $T$ and $\eta$ as BO iterates. The settings are the same as those in Figure \ref{Figure5}.}
\label{Figure2}
\end{figure*}

 The relationship between $\alpha$ and $\eta$ is given in Figure \ref{Figure6}. On \textit{MT-Defect} and \textit{MT}, the optimal region for same DNNs on \textit{MT-Defect} and \textit{MT} illustrates that the similarity of the hyperparameter search space, illustrating the stability of GI-LS when adding perturbed `Free' samples to \textit{MT}. The value of $\eta$ with better performance is small for \textit{ResNet18} and \textit{ResNet50}, while it increases for \textit{ResNet34}. On the contrary, the $\alpha$'s value is large for \textit{ResNet18}, while it gradually decreasing for \textit{ResNet34} and \textit{ResNet50}. As the DNNs size increases, there is an approximately inverse relationship between \(\alpha\) and \(\eta\). For \textit{Wood} and \textit{Carpet}, smaller $\eta$ tends to achieve relatively better performance while $\alpha$ is sensitive to the size of DNNs, as shown in Figure \ref{Figure2}. Since the top-1 accuracy is relatively stable on \textit{Wood} and \textit{Carpet}, given in Figure \ref{Figure1}, the sensitivity of \(\alpha\) for these datasets does not cause a significant impact on DNNs performance. 
\par The relationship of $T$ and $\eta$ with the iteration of BO is shown in Figure \ref{Figure2}. Compared with the results on \textit{MT-Defect} and \textit{MT}, the relationships between the two parameters are relatively similar for the same DNNs, which further illustrates that adding perturbed `Free' samples has few impact on the search space of hyperparameters and the performance of DNNs. Note that  $T$ influences training efficiency, there exists a trade-off between DNNs size and $T$. As the size of the DNNs increases, appropriately decreasing \(T\)'s value can enhance their performance and improve training efficiency somewhat. Regarding \textit{Wood} and \textit{Carpet}, Figure \ref{Figure1} shows that the fluctuations in top-1 accuracy are relatively stable. Therefore, we can select a smaller \(T\) to train DNNs, which maintains performance and improves training efficiency. The results of \textit{ResNet34} demonstrate that a larger $\eta$ can achieve better performance on \textit{MT-Defect} and \textit{MT}, while decreasing $\eta$ tends to achieve better performance when $T$ is decreased. 
\begin{figure}[htbp]
\centering
\includegraphics[width=6.5in]{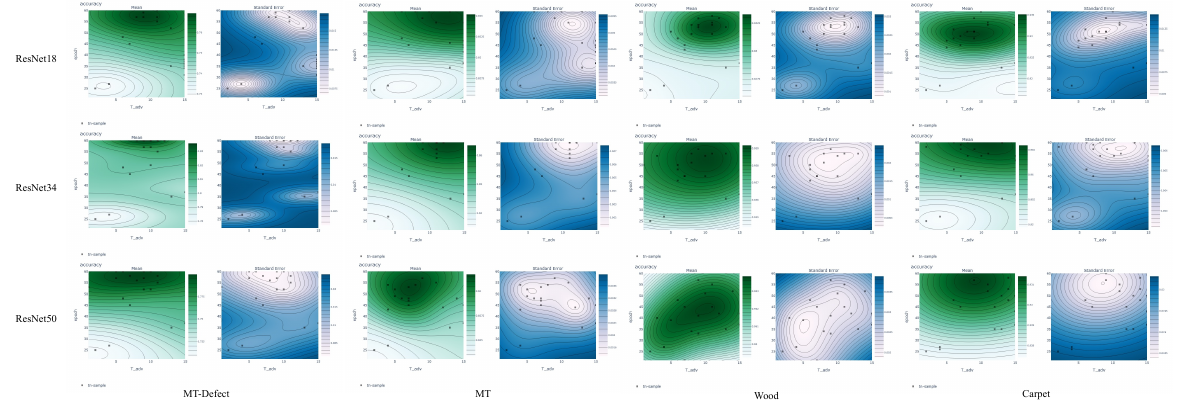}
\caption{The relationship between $T$ and epoch with BO iterates. The settings are the same as those in Figure \ref{Figure5}.}
\label{Figure7}
\end{figure}
\par The relationship between $T$ and epoch is shown in Figure \ref{Figure7}. Clearly, these two parameters relate to the training efficiency of DNNs. The former controls the process for generating data, while the latter influences the training process of DNNs. The results of \textit{MT-Defect} and \textit{MT} demonstrates that the value of epoch decreases appropriately can also achieve better performance with the increasing of $T$. This means that there is a trade-off between the iteration of perturbation for existing data to unseen domains and the iteration of training DNNs parameters. For \textit{ResNet18}, \(T\) values  concentrating around $15$ achieving superior performance. With the increase in DNNs size, the value of \(T\) decreases to some extent. This indicates that the selection of feature extractor influences the perturbations of existing data. Since the performance of DNNs on \textit{Wood} and \textit{Carpet} is relatively stable, decreasing \(T\) and the number of epochs can still maintain good performance of DNNs and improve training efficiency.
\par In practical applications, the experience of selecting hyperparameter combination for GI-LS on small-scale anomaly datasets is summarized as follows: First, we are supposed to select an appropriate feature extractor based on the scale of the datasets. If the dataset accuracy fluctuates significantly during BO iterations, as demonstrated by the results of \textit{ResNet34} on \textit{MT-Defect} and \textit{MT}, a combination of small \(T\) and large \(\alpha\), \(\eta\), and epochs generally benefits model generalization. Otherwise, the hyperparameter selections are flexible. We can focus on improving training efficiency by adjusting down the hyperparameters related to the number of iterations while maintaining the excellent performance of DNNs, as illustrated by the results on \textit{Wood} and \textit{Carpet}.
\subsection{Comparison to Existing Data Augmentation Method}\label{app9}\
In this section, using \textit{MT-Defect} and \textit{MT} as examples, we conduct seven data augmentation methods to compare with the proposed GI-LS method. We regard the DNNs trained with cross-entropy loss without LS and existing images from \textit{MT-Defect} and \textit{MT} as the benchmark. Table \ref{Table4} shows the top-1 accuracy results of Cutmix (\citeauthor{yun2019cutmix}, \citeyear{yun2019cutmix}), Cutout (\citeauthor{devries2017improved}, \citeyear{devries2017improved}), and Mixup (\citeauthor{zhang2018mixup}, \citeyear{zhang2018mixup}), while Table \ref{Table1} records the top-1 accuracy results of PGD (\citeauthor{madry2018towards}, \citeyear{madry2018towards}), FGSM (\citeauthor{goodfellow2014explaining}, \citeyear{goodfellow2014explaining}), and CW attacking (\citeauthor{carlini2017towards}, \citeyear{carlini2017towards}). We set the maximum perturbation $\epsilon = \frac{8}{225}$ for PGD and FGSM. Specifically, we use torchattacks package to achieve adversarial perturbation methods (\citeauthor{kim2020torchattacks}, \citeyear{kim2020torchattacks}).
\begin{table*}[htbp]
\renewcommand\arraystretch{0.7}
\caption{Comparison with the existing data mixing method ($\%$)}
\label{Table4}
\centering
\setlength{\tabcolsep}{5.0mm}{\begin{tabular}{cccccccccccccccc}
\toprule
Dataset &DNNs &Benchmark &GI-LS &Cutmix &Cutout &Mixup  \\
\hline
\bfseries \textit{MT-Defect} &\textit{ResNet18}  &65.2 &\bfseries77.5 &64.0 &64.0 &65.7 \\
                             &\textit{ResNet34}  &66.9 &\bfseries84.3 &59.6 &65.2 &61.2 \\
                             &\textit{ResNet50}  &70.8 &\bfseries79.8 &64.6 &65.7 &67.4 \\
\hline
\bfseries \textit{MT}        &\textit{ResNet18}  &76.2 &\bfseries84.2 &75.9 &74.3 &74.9 \\
                             &\textit{ResNet34}  &75.0 &\bfseries86.6 &75.1 &74.7 &76.0 \\
                             &\textit{ResNet50}  &77.2 &\bfseries84.4 &76.6 &77.2 &77.3 \\
\bottomrule
\end{tabular}}
\end{table*}
\begin{table*}[htbp]
\renewcommand\arraystretch{0.7}
\caption{Comparison with the existing adversarial perturbation method ($\%$)}
\label{Table1}
\centering
\setlength{\tabcolsep}{5.0mm}{\begin{tabular}{cccccccccccccccc}
\toprule
Dataset &DNNs &PGD ($L_{\infty}$) &FGSM ($L_\infty$) &CW($L_\infty$) &PGDL2($L_2$) \\
\hline
\bfseries \textit{MT-Defect} &\textit{ResNet18}  &34.3 &34.3 &46.1 &38.8 \\
&\textit{ResNet34}                               &34.8 &34.8 &37.1 &37.6 \\
&\textit{ResNet50}                               &33.7 &33.2 &43.8 &50.0 \\
\hline
\bfseries \textit{MT} &\textit{ResNet18}         &71.5 &71.6 &72.1 &71.7 \\
&\textit{ResNet34}                               &71.5 &71.8 &72.0 &72.0 \\
&\textit{ResNet50}                               &71.5 &71.8 &72.7 &72.8 \\
\bottomrule
\end{tabular}}
\end{table*}
\par The results of Tables \ref{Table4} and \ref{Table1} demonstrate that GI-LS achieves superior performance compared to the benchmark and the other seven data augmentation methods, confirming the effectiveness of the theoretical analysis in Section \ref{sec:3}. Note that the above seven data methods operate on the original images while GI-LS focus on the feature map corresponding to semantic space, illustrating that the reasonable of defining perturbations on the semantic space in visual space. Compared with the benchmark and LS in Table \ref{Table100}, the performance of DNNs does not improve, indicating that multiple labels for the small-scale data do not benefit the model's generalization. In contrast, GI-LS achieves multiple labels for multiple data and significantly improves model generalization. GI-LS introduces perturbations to existing data to generate new data. Cutmix, Cutout, and Mixup reorganize parts of the images to generate new data. These methods may change the characteristics of the existing data and result in poor generalization. Moreover, from the results of Table \ref{Table4},  we find all performance of DNNs on \textit{MT} improve significantly compared to the those on \textit{MT-Defect}, illustrating the critical role of `Free' samples for improving model's generalization again.  
\par The adversarial samples have a severe negative effect on \textit{MT-Defect}, as illustrated by the results in Table \ref{Table1}. Even so, with the increase in perturbed `Free' samples, the performance of DNNs improves significantly, although they still lag behind the benchmark. This demonstrates that perturbation on `Free' samples can defend against adversarial attacks to some extent. As seen from the above review, adding perturbed `Free' samples benefits the performance of DNNs in GI-LS, illustrating the positive effect of `Free' samples when training DNNs. The adversarial data augmentation methods in Table \ref{Table1} can be summarized as perturbations to the existing data. However, PGD, FGSM, CW, and PGDL2 give imperceptible perturbations to the existing samples, causing DNNs to perform poorly. In contrast, the perturbations of GI-LS are large and easy for models to classify (\citeauthor{volpi2018generalizing}, \citeyear{volpi2018generalizing}). Indeed, GI-LS provides multiple characteristics around the existing data when training DNNs..
\par We next add the adversarial perturbations in Table \ref{Table1} to the samples generated by GI-LS to validate the defend mechanism of GI-LS. Specifically, we still train DNNs using both these perturbed data and the original data. The results are presented in Table \ref{Table5}. Compared with the benchmark in Table \ref{Table4}, we observe that the performance of DNNs still improves significantly, validating the effectiveness of GI-LS under adversarial attack. Since we explore a series of worst-case distributions by considering LS, the shifted data distribution can still reflect the characteristics of the existing data even when influenced by the adversarial attack. Moreover, we see that the performance of DNNs decreases to a certain part compared with GI-LS in most cases, which presents that the adversarial attacks in Table \ref{Table1} for existing data indeed have negative influence on the samples generated by GI-LS.
\begin{table*}[htbp]
\renewcommand\arraystretch{0.7}
\caption{Combination with the existing perturbation method ($\%$)}
\label{Table5}
\centering
\setlength{\tabcolsep}{4.0mm}{\begin{tabular}{cccccccccccccccc}
\toprule
Dataset &Method &GI-LS-PGD &GI-LS-FGSM &GI-LS-CW &GI-LS-PGDL2 \\
\hline
\bfseries \textit{MT-Defect} &\textit{ResNet18} &74.2 &74.2 &74.2 &77.5 \\
&\textit{ResNet34}                              &85.4 &83.2 &83.2 &84.8\\
&\textit{ResNet50}                              &76.4 &71.7 &76.4 &71.5\\
\hline
\bfseries \textit{MT} &\textit{ResNet18}        &82.3 &82.3 &82.3 &82.3\\
&\textit{ResNet34}                              &83.3 &83.3 &83.3 &83.3\\
&\textit{ResNet50}                              &82.9 &82.3 &82.9 &82.9\\
\bottomrule
\end{tabular}}
\end{table*}
\subsection{Experiment for $\gamma$ Perturbation}\label{app10}\
Theorem \ref{t7} provides the bounds of the robust surrogate loss $\boldsymbol\phi_{\boldsymbol\gamma}(\theta ; (z^\prime, y))$, which provide perturbations denoted as $L(\theta)$ on $\gamma$. The top-1 accuracy metrics of DNNs can serve as indicators for the bounds, measuring the distribution of generating samples consistent with $\boldsymbol\phi_{\boldsymbol\gamma}(\theta ; (z^\prime, y))$. In our experiments, we perturb the value of $\gamma$ by simulating diverse magnitudes of $L(\theta)$ in Algorithm \ref{alg:Framwork}. The other hyperparameters are the same as those used in Table \ref{Table101}. Starting with $\gamma=10^{-3}$, we perturbed its denominator by $\pm0.1$, $\pm1$, $\pm10$, and $\pm100$. Detailed findings can be found in Table \ref{Table6}. 
\begin{table*}[!ht]
\renewcommand\arraystretch{0.7}
\caption{Top 1 accuracy on testing data with different $\gamma \pm L(\theta)$ ($\%$)}
\centering  
\setlength{\tabcolsep}{3.5mm}{\begin{tabular}{ccccccccccccccccccccccccc}
\toprule
Dataset &DNNs  &$\frac{1}{1000.1}$ &$\frac{1}{999.9}$ &$\frac{1}{1001}$ &$\frac{1}{999}$ &$\frac{1}{1010}$ &$\frac{1}{990}$ &$\frac{1}{1100}$ &$\frac{1}{900}$\\
\hline
\bfseries \textit{MT-Defect} &\textit{ResNet18}   &77.5 &77.5 &77.5 &77.5 &77.5 &77.5 &77.5 &77.5\\
&\textit{ResNet34}            &84.3 &84.3 &84.3 &84.3 &84.3 &84.3 &84.3 &84.3 \\
&\textit{ResNet50}           &79.8 &79.8 &79.8  &79.8 &79.8 &79.8 &79.8 &79.8\\
    \hline
\bfseries \textit{MT} &\textit{ResNet18}  &84.2 &84.2 &84.2 &84.2 &84.2 &84.2 &84.2 &84.2  \\
&\textit{ResNet34}            &86.6 &86.6 &86.6 &86.6 &86.6 &86.6 &86.6 &86.6  \\
&\textit{ResNet50}            &84.4 &84.4 &84.4 &84.4 &84.4 &84.4 &84.4 &84.4  \\
    \bottomrule
    \end{tabular}}
\label{Table6}
\end{table*}
\par We see that the results remain fairly stable across \textit{ResNet18}, \textit{ResNet34}, and \textit{ResNet50} for both \textit{MT-Defect} and \textit{MT} when perturbing \(\gamma\) with different magnitudes. This implies that the combination of hyperparameters in GI-LS has some robustness to defend against perturbations on \(\gamma\) when shifting data to unseen domains. Moreover, this also illustrates that the generated samples with perturbations are consistent with those without perturbations. In other words, the perturbations of \(\gamma\) with different magnitudes have few influence on the process of generating data.
\section{Conclusion}\label{sec:5}\
In this paper, we propose a two-stage DRO-LS model as a new data augmentation, which considers the regularization effect of LS integrated with DRO to address the issue of overfitting when applying DNNs in few data scenarios. Specifically, we utilize the Wasserstein distance within the DRO framework to construct the ambiguity set and ensure computational feasibility via Lagrangian relaxation. We propose a surrogate loss and prove that it is equivalent to the LS loss and a regularization term. We find that the LS regularization effect can be extended to the regularization term, which contains the parameters of the final fully connected layer of DNNs. Furthermore, we prove that there is a bound for the surrogate loss, which can be used to measure the quality of generating samples consistent with the surrogate loss. Different from the general LS, DRO-LS actually provide multiple characteristics of existing data with multiple labels. To solve DRO-LS model, we develop GI-LS algorithm, corresponding to the two stages of DRO-LS model, to shift existing data and generate new data. We provide the bound of GI-LS, indicating that the perturbations with LS on existing data does not influence the convergence of GI-LS. Finally, we apply GI-LS to various small-scale anomaly cases. Since GI-LS involves a series of hyperparameters to adjust, we apply BO to search for the relatively optimal hyperparameters combination on the cases. We analyze the progress from the viewpoint of classification and data augmentation. Our experimental results show superior performance compared to other general data augmentation methods. 
\par There are some limitations in our research that need to be addressed in future work. As GI-LS contains hyperparameters related to both training efficiency and performance, one of the next steps in our research is to investigate how to better balance efficiency and performance accurately when applying GI-LS. Additionally, the bound of the surrogate loss is intimately connected to the process of generating data. Different magnitudes do not influence this process. Thus, a detailed investigation into how the value of perturbations affects the generation of samples is necessary. Moreover, it is necessary to further investigate the performance of GI-LS beyond merely on small-scale image data.
\ACKNOWLEDGMENT{%
We would like to thank the anonymous reviewers for their insightful suggestions enhancing the quality of this paper.
}

\bibliographystyle{informs2014} 
\bibliography{references.bib} 

\begin{thebibliography}{48}
\providecommand{\natexlab}[1]{#1}
\providecommand{\url}[1]{\texttt{#1}}
\providecommand{\urlprefix}{URL }

\bibitem[{Aghasi et~al.(2024)Aghasi, Rai, \protect\BIBand{} Xia}]{aghasi2024deep}
Aghasi A, Rai A, Xia Y (2024) A deep learning and image processing pipeline for object characterization in firm operations. \emph{INFORMS Journal on Computing} 36(2):616--634.

\bibitem[{AlBahar et~al.(2021)AlBahar, Kim, \protect\BIBand{} Yue}]{albahar2021robust}
AlBahar A, Kim I, Yue X (2021) A robust asymmetric kernel function for bayesian optimization, with application to image defect detection in manufacturing systems. \emph{IEEE Transactions on Automation Science and Engineering} .

\bibitem[{Bai et~al.(2024)Bai, He, Jiang, \protect\BIBand{} Obloj}]{bai2024wasserstein}
Bai X, He G, Jiang Y, Obloj J (2024) Wasserstein distributional robustness of neural networks. \emph{Advances in Neural Information Processing Systems} 36.

\bibitem[{Baird et~al.(2022)Baird, Liu, \protect\BIBand{} Sparks}]{baird2022high}
Baird SG, Liu M, Sparks TD (2022) High-dimensional bayesian optimization of 23 hyperparameters over 100 iterations for an attention-based network to predict materials property: A case study on crabnet using ax platform and saasbo. \emph{Computational Materials Science} 211:111505.

\bibitem[{Bergmann et~al.(2019)Bergmann, Fauser, Sattlegger, \protect\BIBand{} Steger}]{bergmann2019mvtec}
Bergmann P, Fauser M, Sattlegger D, Steger C (2019) Mvtec ad--a comprehensive real-world dataset for unsupervised anomaly detection. \emph{Proceedings of the IEEE/CVF conference on computer vision and pattern recognition}, 9592--9600.

\bibitem[{Blanchet \protect\BIBand{} Murthy(2019)}]{blanchet2019quantifying}
Blanchet J, Murthy K (2019) Quantifying distributional model risk via optimal transport. \emph{Mathematics of Operations Research} 44(2):565--600.

\bibitem[{Blanchet et~al.(2021)Blanchet, Murthy, \protect\BIBand{} Nguyen}]{blanchet2021statistical}
Blanchet J, Murthy K, Nguyen VA (2021) Statistical analysis of wasserstein distributionally robust estimators. \emph{Tutorials in Operations Research: Emerging Optimization Methods and Modeling Techniques with Applications}, 227--254 (INFORMS).

\bibitem[{Bonnans \protect\BIBand{} Shapiro(2013)}]{bonnans2013perturbation}
Bonnans JF, Shapiro A (2013) \emph{Perturbation analysis of optimization problems} (Springer Science \& Business Media).

\bibitem[{Carlini \protect\BIBand{} Wagner(2017)}]{carlini2017towards}
Carlini N, Wagner D (2017) Towards evaluating the robustness of neural networks. \emph{2017 ieee symposium on security and privacy (sp)}, 39--57 (Ieee).

\bibitem[{DeVries \protect\BIBand{} Taylor(2017)}]{devries2017improved}
DeVries T, Taylor GW (2017) Improved regularization of convolutional neural networks with cutout. \emph{arXiv preprint arXiv:1708.04552} .

\bibitem[{Duchi et~al.(2021)Duchi, Glynn, \protect\BIBand{} Namkoong}]{duchi2021statistics}
Duchi JC, Glynn PW, Namkoong H (2021) Statistics of robust optimization: A generalized empirical likelihood approach. \emph{Mathematics of Operations Research} 46(3):946--969.

\bibitem[{Gao et~al.(2017)Gao, Chen, \protect\BIBand{} Kleywegt}]{gao2017distributional}
Gao R, Chen X, Kleywegt AJ (2017) Distributional robustness and regularization in statistical learning. \emph{arXiv preprint arXiv:1712.06050} .

\bibitem[{Garc{\'\i}a~Trillos \protect\BIBand{} Garc{\'\i}a~Trillos(2022)}]{garcia2022regularized}
Garc{\'\i}a~Trillos CA, Garc{\'\i}a~Trillos N (2022) On the regularized risk of distributionally robust learning over deep neural networks. \emph{Research in the Mathematical Sciences} 9(3):54.

\bibitem[{Ghadimi \protect\BIBand{} Lan(2013)}]{ghadimi2013stochastic}
Ghadimi S, Lan G (2013) Stochastic first-and zeroth-order methods for nonconvex stochastic programming. \emph{SIAM journal on optimization} 23(4):2341--2368.

\bibitem[{Goodfellow et~al.(2014)Goodfellow, Shlens, \protect\BIBand{} Szegedy}]{goodfellow2014explaining}
Goodfellow IJ, Shlens J, Szegedy C (2014) Explaining and harnessing adversarial examples. \emph{arXiv preprint arXiv:1412.6572} .

\bibitem[{He et~al.(2016)He, Zhang, Ren, \protect\BIBand{} Sun}]{he2016deep}
He K, Zhang X, Ren S, Sun J (2016) Deep residual learning for image recognition. \emph{Proceedings of the IEEE conference on computer vision and pattern recognition}, 770--778.

\bibitem[{Huang et~al.(2020)Huang, Qiu, \protect\BIBand{} Yuan}]{huang2020surface}
Huang Y, Qiu C, Yuan K (2020) Surface defect saliency of magnetic tile. \emph{The Visual Computer} 36(1):85--96.

\bibitem[{Jiao et~al.(2022)Jiao, Yang, \protect\BIBand{} Song}]{jiao2022distributed}
Jiao Y, Yang K, Song D (2022) Distributed distributionally robust optimization with non-convex objectives. \emph{Advances in neural information processing systems} 35:7987--7999.

\bibitem[{Kim(2020)}]{kim2020torchattacks}
Kim H (2020) Torchattacks: A pytorch repository for adversarial attacks. \emph{arXiv preprint arXiv:2010.01950} .

\bibitem[{Kuhn et~al.(2019)Kuhn, Esfahani, Nguyen, \protect\BIBand{} Shafieezadeh-Abadeh}]{kuhn2019Wasserstei}
Kuhn D, Esfahani PM, Nguyen VA, Shafieezadeh-Abadeh S (2019) Wasserstein distributionally robust optimization: Theory and applications in machine learning. \emph{Operations research \& management science in the age of analytics}, 130--166 (Informs).

\bibitem[{Levy et~al.(2020)Levy, Carmon, Duchi, \protect\BIBand{} Sidford}]{levy2020large}
Levy D, Carmon Y, Duchi JC, Sidford A (2020) Large-scale methods for distributionally robust optimization. \emph{Advances in Neural Information Processing Systems} 33:8847--8860.

\bibitem[{Li et~al.(2020)Li, Dasarathy, \protect\BIBand{} Berisha}]{li2020regularization}
Li W, Dasarathy G, Berisha V (2020) Regularization via structural label smoothing. \emph{International Conference on Artificial Intelligence and Statistics}, 1453--1463 (PMLR).

\bibitem[{Liu et~al.(2022)Liu, Wu, Li, \protect\BIBand{} Cui}]{liu2022distributionally}
Liu J, Wu J, Li B, Cui P (2022) Distributionally robust optimization with data geometry. \emph{Advances in neural information processing systems} 35:33689--33701.

\bibitem[{Lukasik et~al.(2020)Lukasik, Bhojanapalli, Menon, \protect\BIBand{} Kumar}]{lukasik2020does}
Lukasik M, Bhojanapalli S, Menon A, Kumar S (2020) Does label smoothing mitigate label noise? \emph{International Conference on Machine Learning}, 6448--6458 (PMLR).

\bibitem[{Luo et~al.(2020)Luo, Mobahi, \protect\BIBand{} Bengio}]{luo2020data}
Luo C, Mobahi H, Bengio S (2020) Data augmentation via structured adversarial perturbations.

\bibitem[{Madry et~al.(2018)Madry, Makelov, Schmidt, Tsipras, \protect\BIBand{} Vladu}]{madry2018towards}
Madry A, Makelov A, Schmidt L, Tsipras D, Vladu A (2018) Towards deep learning models resistant to adversarial attacks. \emph{International Conference on Learning Representations}.

\bibitem[{M{\"u}ller et~al.(2019)M{\"u}ller, Kornblith, \protect\BIBand{} Hinton}]{muller2019does}
M{\"u}ller R, Kornblith S, Hinton GE (2019) When does label smoothing help? \emph{Advances in neural information processing systems} 32.

\bibitem[{Noyan et~al.(2022)Noyan, Rudolf, \protect\BIBand{} Lejeune}]{noyan2022distributionally}
Noyan N, Rudolf G, Lejeune M (2022) Distributionally robust optimization under a decision-dependent ambiguity set with applications to machine scheduling and humanitarian logistics. \emph{INFORMS Journal on Computing} 34(2):729--751.

\bibitem[{Oberman \protect\BIBand{} Calder(2019)}]{oberman2019lipschitz}
Oberman AM, Calder J (2019) Lipschitz regularized deep neural networks generalize.

\bibitem[{Peck et~al.(2023)Peck, Goossens, \protect\BIBand{} Saeys}]{peck2023introduction}
Peck J, Goossens B, Saeys Y (2023) An introduction to adversarially robust deep learning. \emph{IEEE Transactions on Pattern Analysis and Machine Intelligence} .

\bibitem[{Rockafellar \protect\BIBand{} Wets(2009)}]{rockafellar2009variational}
Rockafellar RT, Wets RJB (2009) \emph{Variational analysis}, volume 317 (Springer Science \& Business Media).

\bibitem[{Rychener et~al.(2023)Rychener, Kuhn, \protect\BIBand{} Sutter}]{rychener2023end}
Rychener Y, Kuhn D, Sutter T (2023) End-to-end learning for stochastic optimization: A bayesian perspective. \emph{International Conference on Machine Learning}, 29455--29472 (PMLR).

\bibitem[{Semelhago et~al.(2021)Semelhago, Nelson, Song, \protect\BIBand{} W{\"a}chter}]{semelhago2021rapid}
Semelhago M, Nelson BL, Song E, W{\"a}chter A (2021) Rapid discrete optimization via simulation with gaussian markov random fields. \emph{INFORMS Journal on Computing} 33(3):915--930.

\bibitem[{Shafieezadeh-Abadeh et~al.(2019)Shafieezadeh-Abadeh, Kuhn, \protect\BIBand{} Esfahani}]{shafieezadeh2019regularization}
Shafieezadeh-Abadeh S, Kuhn D, Esfahani PM (2019) Regularization via mass transportation. \emph{Journal of Machine Learning Research} 20(103):1--68.

\bibitem[{Shafieezadeh~Abadeh et~al.(2015)Shafieezadeh~Abadeh, Mohajerin~Esfahani, \protect\BIBand{} Kuhn}]{shafieezadeh2015distributionally}
Shafieezadeh~Abadeh S, Mohajerin~Esfahani PM, Kuhn D (2015) Distributionally robust logistic regression. \emph{Advances in Neural Information Processing Systems} 28.

\bibitem[{Shu \protect\BIBand{} Song(2014)}]{shu2014dynamic}
Shu J, Song M (2014) Dynamic container deployment: two-stage robust model, complexity, and computational results. \emph{INFORMS Journal on Computing} 26(1):135--149.

\bibitem[{Sinha et~al.(2018)Sinha, Namkoong, \protect\BIBand{} Duchi}]{sinha2018certifiable}
Sinha A, Namkoong H, Duchi J (2018) Certifiable distributional robustness with principled adversarial training. \emph{International Conference on Learning Representations}.

\bibitem[{Szegedy et~al.(2016)Szegedy, Vanhoucke, Ioffe, Shlens, \protect\BIBand{} Wojna}]{szegedy2016rethinking}
Szegedy C, Vanhoucke V, Ioffe S, Shlens J, Wojna Z (2016) Rethinking the inception architecture for computer vision. \emph{Proceedings of the IEEE conference on computer vision and pattern recognition}, 2818--2826.

\bibitem[{Tabernik et~al.(2020)Tabernik, {\v{S}}ela, Skvar{\v{c}}, \protect\BIBand{} Sko{\v{c}}aj}]{tabernik2020segmentation}
Tabernik D, {\v{S}}ela S, Skvar{\v{c}} J, Sko{\v{c}}aj D (2020) Segmentation-based deep-learning approach for surface-defect detection. \emph{Journal of Intelligent Manufacturing} 31(3):759--776.

\bibitem[{Villani(2021)}]{villani2021topics}
Villani C (2021) \emph{Topics in optimal transportation}, volume~58 (American Mathematical Soc.).

\bibitem[{Volpi et~al.(2018)Volpi, Namkoong, Sener, Duchi, Murino, \protect\BIBand{} Savarese}]{volpi2018generalizing}
Volpi R, Namkoong H, Sener O, Duchi JC, Murino V, Savarese S (2018) Generalizing to unseen domains via adversarial data augmentation. \emph{Advances in neural information processing systems} 31.

\bibitem[{Wah et~al.(2011)Wah, Branson, Welinder, Perona, \protect\BIBand{} Belongie}]{wah2011caltech}
Wah C, Branson S, Welinder P, Perona P, Belongie S (2011) The caltech-ucsd birds-200-2011 dataset .

\bibitem[{Wang et~al.(2021)Wang, Huang, Song, Pan, Xia, \protect\BIBand{} Wu}]{wang2021regularizing}
Wang Y, Huang G, Song S, Pan X, Xia Y, Wu C (2021) Regularizing deep networks with semantic data augmentation. \emph{IEEE Transactions on Pattern Analysis and Machine Intelligence} 44(7):3733--3748.

\bibitem[{Xu et~al.(2023)Xu, Jiang, Svetozarevic, \protect\BIBand{} Jones}]{xu2023constrained}
Xu W, Jiang Y, Svetozarevic B, Jones C (2023) Constrained efficient global optimization of expensive black-box functions. \emph{International Conference on Machine Learning}, 38485--38498 (PMLR).

\bibitem[{Yuan et~al.(2020)Yuan, Tay, Li, Wang, \protect\BIBand{} Feng}]{yuan2020revisiting}
Yuan L, Tay FE, Li G, Wang T, Feng J (2020) Revisiting knowledge distillation via label smoothing regularization. \emph{Proceedings of the IEEE/CVF Conference on Computer Vision and Pattern Recognition}, 3903--3911.

\bibitem[{Yun et~al.(2019)Yun, Han, Oh, Chun, Choe, \protect\BIBand{} Yoo}]{yun2019cutmix}
Yun S, Han D, Oh SJ, Chun S, Choe J, Yoo Y (2019) Cutmix: Regularization strategy to train strong classifiers with localizable features. \emph{Proceedings of the IEEE/CVF international conference on computer vision}, 6023--6032.

\bibitem[{Zhang et~al.(2018)Zhang, Cisse, Dauphin, \protect\BIBand{} Lopez-Paz}]{zhang2018mixup}
Zhang H, Cisse M, Dauphin YN, Lopez-Paz D (2018) mixup: Beyond empirical risk minimization. \emph{International Conference on Learning Representations}.

\bibitem[{Zhou et~al.(2022)Zhou, Liu, Qiao, Xiang, \protect\BIBand{} Loy}]{zhou2022domain}
Zhou K, Liu Z, Qiao Y, Xiang T, Loy CC (2022) Domain generalization: A survey. \emph{IEEE Transactions on Pattern Analysis and Machine Intelligence} 45(4):4396--4415.

\end{thebibliography}
\begin{APPENDIX}\
\section{Appendix}
\subsection{Proof of Theorem \ref{t1}}\label{App1}\
Our proof is based on \citeauthor{sinha2018certifiable}(\citeyear{sinha2018certifiable}). Notice that $\boldsymbol\ell(\theta; (Z, Y))$ is continuity for $Z$ and $W_\theta(P, Q)$ satisfies the closed convexity. If $P = Q$, then $W_\theta(P, Q) = 0$. Therefore, Slater's condition holds. We apply the duality results (\citeauthor{blanchet2019quantifying}, \citeyear{blanchet2019quantifying}) and obtain
\begin{equation}
    \sup _{P: W_\theta(P, Q) \leq \rho} \int \boldsymbol{\ell}(\theta; (z, y)) d P(z) = \inf _{\gamma \geq 0}\int \boldsymbol{\ell}(\theta; (z, y)) d P(z) -\gamma W_\theta(P, Q) + \gamma \rho
    \label{a1}
    \tag{1}
\end{equation}
\par For any $M \in \Pi(P, Q)$, we have: $\int \boldsymbol{\ell}(\theta; (z, y)) d P(z) = \int \boldsymbol{\ell}(\theta; (z, y)) d M(z, z^\prime)$. The right-hand of the equality (\ref{a1}) can be reformulated as follows.
\begin{equation*}
\begin{aligned}
    &\inf_{\gamma \geq 0}\int \boldsymbol{\ell}(\theta; (z, y)) d P(z) -\gamma W_\theta(P, Q) \\ & = \int \boldsymbol{\ell}(\theta; (z, y)) dM(z, z^\prime) - \inf_{M \in \Pi(P, Q)} \int \gamma c_\theta(z, z^\prime)dM(z, z^\prime) \\
    & = \sup_{M \in \Pi(P, Q)}\int \boldsymbol\ell(\theta; (z, y))-\gamma c_\theta(z, z^\prime)dM(z, z^\prime)\\
\end{aligned}
\end{equation*}
\par We reformulate equation (\ref{a1}), and get
\begin{equation}
    \sup _{P: W_\theta(P, Q) \leq \rho} \int \boldsymbol{\ell}(\theta; (z, y)) d P(z) = \inf_{\gamma \geq 0}\sup_{M \in \Pi(P, Q)}\int \boldsymbol\ell(\theta; (z,y))-\gamma c_\theta(z, z^\prime)dM(z, z^\prime) + \gamma\rho
    \label{a2}
    \tag{2}
\end{equation}
\par Using Jensen's inequality, the right-hand of equality (\ref{a2}) is reformulated as follows.
\begin{equation*}
\begin{aligned}
    \sup_{M \in \Pi(P, Q)}\int \boldsymbol\ell(\theta; (z,y))-\gamma c_\theta(z, z^\prime)dM(z, z^\prime) & = \sup_{M \in \Pi(P, Q)}\int \boldsymbol\ell(\theta; (z,y))-\gamma c_\theta(z, z^\prime)dQ(z^\prime)\\
    & \leq \int \sup_{z}[\boldsymbol\ell(\theta; (z,y))-\gamma c_\theta(z, z^\prime)]dQ(z^\prime)
\end{aligned}
\end{equation*}
\par We define the space of regular conditional probability space $\mathcal{B}$ from $Z^\prime$ to $Z$, where $\mathcal{B} := P(A | z^\prime)$ for each $z^\prime$ and each measurable set $A$. Then, we get
\begin{equation*}
    \sup_{M \in \Pi(P, Q)}\int \boldsymbol\ell(\theta; (z, y))-\gamma c_\theta(z, z^\prime)dM(z, z^\prime) \geq \sup_{P \in \mathcal{B}}\int \boldsymbol\ell(\theta; (z,y))-\gamma c_\theta(z, z^\prime)dP(z|z^\prime)dQ(z^\prime)
\end{equation*}
\par Since $\mathcal{B}$ is the regular conditional probability space, the mapping $z^\prime \mapsto P(A | z^\prime)$ is measurable for each $A$. Let $\mathcal{Z}$ denote the space of all measurable mappings $z^\prime \mapsto z(z^\prime)$ from $Z^\prime$ to $Z$. Using the Theorem 14.60 of the \citeauthor{rockafellar2009variational}(\citeyear{rockafellar2009variational}), we have
\begin{equation}
    \sup_{z \in Z}\int [\boldsymbol\ell(\theta; (z(z^\prime), y))-\gamma c_\theta(z(z^\prime), z^\prime)]dQ(z^\prime) = \int\sup_{z \in Z}[\boldsymbol\ell(\theta; (z, y))-\gamma c_\theta(z, z^\prime)]dQ(z^\prime)
    \label{a3}
    \tag{3}
\end{equation}
\par Observe that $\boldsymbol\ell(\theta; (z,y))-\gamma c_\theta(z, z^\prime)$ is continuous and $c_\theta(z, z^\prime)$ is measurable. Now, let $z(z^\prime)$ be any measurable function that is $\epsilon$-close to obtain the supremum of equality (\ref{a3}). We get
\begin{equation*}
\begin{aligned}
\sup_{P \in \mathcal{B}}\int \boldsymbol\ell(\theta; (z, y))-\gamma c_\theta(z, z^\prime)dP(z|z^\prime ) d Q(z^\prime) & = \sup_{P \in \mathcal{B}}\int \boldsymbol\ell(\theta; (z(z^\prime), y))-\gamma c_\theta(z(z^\prime), z^\prime)d Q(z^\prime) \\
& \geq \int \sup_{z \in Z}[\boldsymbol\ell(\theta; (z, y))-\gamma c_\theta(z, z^\prime)d Q(z^\prime)] - \epsilon \\
& \geq \sup_{M \in \Pi(P, Q)}\int \boldsymbol\ell(\theta; (z,y))-\gamma c_\theta(z, z^\prime)dQ(z^\prime) - \epsilon
\end{aligned}
\end{equation*}
\par Since $\epsilon > 0$ is arbitrary, we have
\begin{equation}
\sup_{M \in \Pi(P, Q)}\int \boldsymbol\ell(\theta; (z,y))-\gamma c_\theta(z, z^\prime)dM(z, z^\prime) = \int\sup_{z \in Z}[\boldsymbol\ell(\theta; (z, y))-\gamma c_\theta(z, z^\prime)]dQ(z^\prime)
\label{a4}
\tag{4}
\end{equation}
\subsection{Proof of Theorem \ref{t2}}\label{App2}\
The cross entropy loss integrated LS for generated data is given as follows.
\begin{equation}
\boldsymbol\ell(\theta ; (z, y)) = -\sum_{j=1}^{k}y_{i}\log p_{j}(\theta, z)
\label{a5}
\tag{5}
\end{equation}
where $y_{i}$ is:
\begin{equation}
y_i=\left\{
\begin{aligned}
&1-\alpha, \qquad& &\text{if $z$ belongs to class $i$,} \\
&\frac{\alpha}{k-1}  \qquad& \  & \text{otherwise.}
\end{aligned}
\label{a6}
\tag{6}
\right.
\end{equation}
\par Using Taylor theorem, we obtain
\begin{equation}
\begin{aligned}
\boldsymbol\ell\left(\theta ;\left(z, y\right)\right) = \boldsymbol\ell\left(\theta ;\left(z^\prime, y\right)\right) + \nabla_{z^\prime}\boldsymbol\ell\left(\theta ;\left(z^\prime, y\right)\right)^T(z-z^{\prime}) + O(\|z - z^\prime\|_2^2).   
\end{aligned}
\label{a7}
\tag{7}
\end{equation}
\par Given that \(O(\|z - z^\prime\|_2^2)\) represents a higher-order term, we replace it with \(\frac{H}{2}\|z - z^\prime\|_2^2\), where \(H\) is a non-identity Hessian matrix. Using equality (\ref{a7}), the robust surrogate loss $\boldsymbol\phi_{\gamma}(\theta ;(z^\prime, y))$ can be reformulated as follows.
\begin{equation}
\begin{aligned}
\boldsymbol{\phi_{\gamma}}(\theta ;(z^\prime, y))=\sup _{z \in \mathcal{Z}}\{\boldsymbol\ell\left(\theta ;\left(z^\prime, y\right)\right) + \nabla_{z^\prime}\boldsymbol\ell\left(\theta ;\left(z^\prime, y\right)\right)^T(z-z^{\prime}) + \frac{H-\gamma I}{2}\|z - z^\prime\|_2^2\}.
\end{aligned}
\label{a8}
\tag{8}
\end{equation}
\par We derive $z$ in equality (\ref{a8}) and obtain
\begin{equation*}
    \nabla_{z^\prime}\boldsymbol\ell\left(\theta ;\left(z^\prime, y\right)\right) + (H - \gamma I)(z - z^\prime) = 0
\end{equation*}
Then, we reach
\begin{equation}
\left\|z^{\prime\star} - z^\prime\right\| = \left\|(H - \gamma I)^{-1} \nabla_{z^{\prime}}\boldsymbol\ell(\theta ;(z^{\prime}, y))\right\|
\label{a9}
\tag{9}
\end{equation}
\par We abuse notation by $f(\theta_r, z^\prime) = f(\theta_r, x)$. Then, we derive $z^\prime$ in $\boldsymbol\ell(\theta;(z^\prime, y))$ to further reformulate equality (\ref{a9}). Using the chain rule, we obtain
\begin{equation*}
\frac{\partial \boldsymbol\ell}{\partial \boldsymbol{z^{\prime}}}=\frac{\partial \boldsymbol\ell}{\partial \boldsymbol{p(\theta, z^{\prime})}} \cdot \frac{\partial \boldsymbol{p(\theta, z^{\prime})}}{\partial \boldsymbol{z^{\prime}}}
\end{equation*}
\begin{equation*}
\frac{\partial \boldsymbol\ell}{\partial {p_{i}(\theta, z^{\prime})}} = \left\{
\begin{aligned}
&-\frac{1-\alpha}{p_{i}(\theta, z^{\prime})},\qquad& \text{if $z^\prime$ belongs to class $i$,} \\
&-\frac{\alpha}{(k-1) \cdot p_{i}(\theta, z^{\prime})} \qquad& \text{otherwise.}
\end{aligned}
\right.
\end{equation*}
\begin{equation*}
\frac{\partial \ell}{\partial \boldsymbol{p(\theta, z^{\prime})}} = \left[-\frac{\alpha}{(k-1) \cdot p_{1}(\theta, z^{\prime})},-\frac{\alpha}{(k-1) \cdot p_{2}(\theta, z^{\prime})}, \ldots,-\frac{1-\alpha}{p_{i}(\theta, z^{\prime})}, \ldots, -\frac{\alpha}{(k-1) \cdot p_{k}(\theta, z^{\prime})}\right]
\end{equation*}
\\
\begin{equation*}
\frac{\partial {p_{i}(\theta, z^{\prime})}}{\partial \boldsymbol{z^{\prime}}} = \left\{
\begin{aligned}
&\frac{\exp \left(\theta_{f, i}^{\top} f\left(\theta_{r}; z^{\prime}\right)\right) \cdot \theta_{f, i}^{\top}\left(\sum_{j=1}^{k} \exp \left(\theta_{f, j}^{\top} f\left(\theta_{r};z^{\prime}\right)\right)-\exp \left(\theta_{f, i}^{\top} f\left(\theta_{r} ; z^{\prime}\right)\right)\right)}{\left[\sum_{j=1}^{k} \exp \left(\theta_{f, j}^{\top} f\left(\theta_{r} ; z^{\prime}\right)\right)\right]^{2}},\ &if\ i=j, \\
\\
&\frac{\theta_{f, j}^{\top} \exp \left(\theta_{f, i}^{\top} f\left(\theta_{r} ; z^{\prime}\right)\right) \cdot \exp \left(\theta_{f, j}^{\top} f\left(\theta_{r} ; z^{\prime}\right)\right)}{\left[\sum_{j=1}^{k} \exp \left(\theta_{f, j}^{\top} f\left(\theta_{r} ; z^{\prime}\right)\right]^{2}\right.} &if\ i \neq j.
\end{aligned}
\right.
\end{equation*}
\par Note that the gradient as a vector, when $i$ is the true class of the sample, we get
\begin{equation*}
\frac{\partial \boldsymbol\ell}{\partial \boldsymbol{z^{\prime}}} = \boldsymbol[(1-\alpha) \cdot p_{1}(\theta ; z^{\prime}) \cdot \theta_{f, 1}^{\top}, (1-\alpha) \cdot p_{2}(\theta ; z^{\prime}) \cdot \theta_{f, 2}^{\top}, \ldots,-(1-\alpha) \cdot\left(\theta_{f, i}^{\top}-p_{i}(\theta ; z^{\prime}) \cdot \theta_{f, i}^{\top}\right),  
\end{equation*}
\begin{equation}
\ldots, (1-\alpha) \cdot p_{k}(\theta ; z) \cdot \theta_{f, k}^{\top}\boldsymbol]
\label{a10}
\tag{10}
\end{equation}
\begin{equation}
\nabla_{z^{\prime}} \boldsymbol\ell(\theta ;(z^{\prime}, y)) = (1-\alpha)\left(\theta_{f, i}^{\top}-\sum_{j=1}^{k} p_{j}(\theta ; z^{\prime}) \cdot \theta_{f, j}^{\top}\right)
\label{a11}
\tag{11}
\end{equation}
\par The $\frac{\partial \boldsymbol\ell}{\partial \boldsymbol{z^{\prime}}}$ of the remaining $k-1$ classes is given as follows.
\begin{equation*}
\begin{split}
\frac{\partial \boldsymbol\ell}{\partial \boldsymbol{z^{\prime}}} = \boldsymbol[\frac{\alpha}{(k-1)} \cdot p_{1}(\theta ; z^{\prime}) \cdot \theta_{f, 1}^{\top}, \frac{\alpha}{(k-1)} \cdot p_{2}(\theta ; z^{\prime}) \cdot \theta_{f, 2}^{\top}, \ldots,-\frac{\alpha}{(k-1)} \cdot\left(\theta_{f, j}^{\top}-p_{j}(\theta ; z^{\prime}) \cdot \theta_{f, j}^{\top}\right), 
\end{split}
\end{equation*}
\begin{equation}
\ldots,\frac{\alpha}{(k-1)} \cdot p_{k}(\theta ; z^{\prime}) \cdot \theta_{f, k}^{\top}\boldsymbol]
\label{a12}
\tag{12}
\end{equation}
\begin{equation}
\nabla_{z^{\prime}} \boldsymbol\ell(\theta ;(z^{\prime}, y)) = \left(\frac{\alpha}{(k-1)}\theta_{f, j}^{\top}-\frac{\alpha}{(k-1)}\sum_{j=1}^{k} p_{j}\left(\theta ; z^{\prime}\right) \cdot \theta_{f, j}^{\top}\right) \qquad j \neq i 
\label{a13}
\tag{13}
\end{equation}
\par Using equality (\ref{a11}) and (\ref{a13}), we get
\begin{equation*}
\begin{split}
\nabla_{z^{\prime}} \boldsymbol\ell(\theta ;(z^{\prime}, y)) =  (1-\alpha)\left(\theta_{f, i}^{\top}-\sum_{j=1}^{k} p_{j}(\theta ; z^{\prime}) \cdot \theta_{f, j}^{\top}\right)
+ \sum_{j\neq i}^{k}\frac{\alpha}{(k-1)}\theta_{f, j}^{\top}-\alpha \sum_{j=1}^{k} p_{j}\left(\theta ; z^{\prime}\right) \cdot \theta_{f, j}^{\top}
\end{split}
\end{equation*}
\begin{equation}
\nabla_{z^\prime} \boldsymbol\ell(\theta ;(z^\prime, y)) = \left((1-\alpha)\theta_{f, i}^{\top} + \frac{\alpha}{(k-1)}\sum_{j\neq i}^{k}\theta_{f, j}^{\top}-\sum_{j=1}^{k} p_{j}\left(\theta ; z^\prime\right) \cdot \theta_{f, j}^{\top}\right)
\label{a14}
\tag{14}
\end{equation}
\subsection{Proof of Theorem \ref{t6}}\label{App5}\
Considering \( f_0(z^{\prime\star}) \) and \( f_1(z^\prime) \), and based on the inequality \(\left\|a+b\right\|_{2} \leq \left\|a\right\|_{2} + \left\|b\right\|_{2}\), we obtain
\begin{equation*}
\begin{aligned}
\left\|f_{0}(z^{\prime\star})-f_{1}(z^\prime)\right\|_{2}&= \left\|\boldsymbol\ell\left(\theta ;\left(z^{\prime\star}, y\right)\right)-\boldsymbol\ell_{1}\left(\theta ;\left(z^\prime, y\right)\right)\right\|_{2}\\
&\leq \left\|\boldsymbol\ell\left(\theta ;\left(z^{\prime\star}, y\right)\right) - \boldsymbol\ell\left(\theta ;\left(z^{\prime}, y\right)\right)\right\|_{2}+ \left\|\boldsymbol\ell_{1}\left(\theta ;\left(z^{\prime\star}, y\right)\right)-\boldsymbol\ell_{1}\left(\theta ;\left(z^{\prime}, y\right)\right) \right\|_{2} \\
&\leq 2L_{0}\left\|z^{\prime\star}-z^{\prime}\right\|_{2}.
\end{aligned}
\end{equation*}
Thus, $f_1 - f_0$ is $2L_0$-Lipschitz.
Using $f_0$, we get
\begin{equation*}
\begin{aligned}
f_{0}(z^{\prime\star}) - f_{0}(z^{\prime})&= f_{1}(z^{\prime\star})+f_{0}(z^{\prime\star})-f_{1}(z^{\prime})-f_{0}(z^{\prime})-f_{1}(z^{\prime\star})+f_{1}(z^{\prime})\\
&\leq 2\left\|f_{1}(z^{\prime\star})-f_{1}(z^\prime)\right\|_{2}+\left\|f_{0}(z^{\prime\star})-f_{0}(z^{\prime})\right\|_{2}\\
&\leq 2\left\|\boldsymbol\ell\left(\theta ;\left(z^{\prime\star}, y\right)\right) - \boldsymbol\ell\left(\theta ;\left(z^{\prime}, y\right)\right)\right\|_{2}+ \left\|\boldsymbol\ell_{1}\left(\theta ;\left(z^{\prime\star}, y\right)\right)-\boldsymbol\ell_{1}\left(\theta ;\left(z^{\prime}, y\right)\right) \right\|_{2}\\
&\leq 3L_{0}\left\|z^{\prime\star}-z^{\prime}\right\|_{2}
\end{aligned}
\end{equation*}
\par According to Lemma \ref{l5}, we have
\begin{equation*}
f_{0}(z^{\prime\star})-f_{0}(z^{\prime}) \geq \gamma\left\|z^{\prime\star}-z^{\prime}\right\|_{2}^{2}
\end{equation*}
\par Then, we obtain
\begin{equation*}
\begin{aligned}
\gamma\left\|z^{\prime\star}-z^{\prime}\right\|_{2}^{2}\leq3L_{0}\left\|z^{\prime\star}-z^{\prime}\right\|_{2}\\
\end{aligned}
\end{equation*}
\begin{equation}
\begin{aligned}
\left\|z^{\prime\star}-z^{\prime}\right\|_{2} \leq \frac{3L_0}{\gamma}
\label{a15}
\end{aligned}
\tag{15}
\end{equation}
\subsection{Proof of Theorem \ref{t7}}\label{App4}\
We consider LS is the sum of $\ell(\theta ;(z, y))$. Using Taylor theorem, we have
\begin{equation}
\left\|\ell\left(\theta ;\left(z, y\right)\right)-\ell(\theta ;(z^{\prime}, y))-\nabla_{z^{\prime}} \ell(\theta ;(z^{\prime}, y))^{\top}\left(z-z^{\prime}\right)\right\|_2 \leq \frac{L}{2} \left\|z-z^{\prime}\right\|_{2}^{2}
\label{a16}
\tag{16}
\end{equation}
\par The upper bound is
\begin{equation*}
\begin{aligned}
\phi_{\gamma}(\theta ;(z^\prime, y)) & \leq  \sup _{z}\left\{\ell(\theta ;(z^{\prime}, y))+\nabla_{z^{\prime}} \ell(\theta ;(z^{\prime}, y))^{\top}\left(z-z^{\prime}\right) -\frac{(\gamma-L)}{2}\left\|z-z^{\prime}\right\|_{2}^{2}\right\} \\
& \leq\ell(\theta ;(z^{\prime}, y))+\frac{1}{2(\gamma-L)}\left\|\nabla_{z^{\prime}} \ell(\theta ;(z^{\prime}, y))\right\|_{2}^{2}
\end{aligned}
\end{equation*}
When $i$ is the true class of sample, we have
\begin{equation}
\phi_{\gamma}(\theta ;(z^\prime, y)) \leq \ell(\theta ;(z^{\prime}, y))+\frac{1-\alpha}{2(\gamma-L)}\left\|\theta_{f, i}^{\top}-\sum_{j=1}^{k} p_{j}(\theta ; z^{\prime}) \cdot \theta_{f, j}^{\top}\right\|_{2}^{2}
\label{a17}
\tag{17}
\end{equation}
\par Similarly, the upper bound of the remaining $k-1$ classes is
\begin{equation}
\phi_{\gamma}(\theta ;(z^\prime, y)) \leq \ell(\theta ;(z^{\prime}, y))+\frac{\alpha}{2(k-1)(\gamma-L)}\left\|\theta_{f, j}^{\top}-\sum_{j=1}^{k} p_{j}(\theta ; z^{\prime}) \cdot \theta_{f, j}^{\top}\right\|_{2}^{2} \qquad j \neq i
\label{a18}
\tag{18}
\end{equation}
\par Using inequalities (\ref{a17}) and (\ref{a18}), we obtain the upper bound of $\boldsymbol\phi_{\boldsymbol\gamma}(\theta ;(z^\prime, y)) - \boldsymbol\ell(\theta ;(z^{\prime}, y))$
\begin{equation*}
\begin{split}
\boldsymbol\phi_{\boldsymbol\gamma}(\theta ; (z^\prime, y)) \leq
\boldsymbol\ell(\theta ;(z^{\prime}, y)) + \|\frac{1-\alpha}{2(\gamma-L)}\theta_{f, i}^{\top}-\frac{1-\alpha}{2(\gamma-L)}\sum_{j=1}^{k} p_{j}(\theta ; z^{\prime}) \cdot \theta_{f, j}^{\top}\\
+\frac{\alpha}{2(k-1)(\gamma-L)}\sum_{j\neq i}^{k}\theta_{f, j}^{\top}-\frac{\alpha}{2(k-1)(\gamma-L)}\sum_{j=1}^{k} p_{j}(\theta ; z^{\prime}) \cdot \theta_{f, j}^{\top}\|_{2}^{2}\\
\end{split}
\end{equation*}
\begin{equation}
\boldsymbol\phi_{\boldsymbol\gamma}(\theta ; (z^\prime, y))-\boldsymbol\ell(\theta ;(z^{\prime}, y)) \leq
 \frac{1}{2(\gamma-L)}\|(1-\alpha)\theta_{f, i}^{\top}
+\frac{\alpha}{(k-1)}\sum_{j\neq i}^{k}\theta_{f, j}^{\top}-\sum_{j=1}^{k} p_{j}(\theta ; z^{\prime}) \cdot \theta_{f, j}^{\top}\|_{2}^{2}\\
\label{a19}
\tag{19}
\end{equation}
\par Similarly, the lower bound of $\boldsymbol\phi_{\boldsymbol\gamma}(\theta ; (z^\prime, y))-\boldsymbol\ell(\theta ;(z^{\prime}, y))$ is
\begin{equation}
\boldsymbol\phi_{\boldsymbol\gamma}(\theta ; (z^\prime, y))-\boldsymbol\ell(\theta ;(z^{\prime}, y)) \geq
 \frac{1}{2(\gamma+L)}\|(1-\alpha)\theta_{f, i}^{\top}
+\frac{\alpha}{(k-1)}\sum_{j\neq i}^{k}\theta_{f, j}^{\top}-\sum_{j=1}^{k} p_{j}(\theta ; z^{\prime}) \cdot \theta_{f, j}^{\top}\|_{2}^{2}\\
\label{a20}
\tag{20}
\end{equation}
\par According to Theorem \ref{t2}, we have
\begin{equation*}
\left\|\nabla_{z^{\prime}} \boldsymbol\ell(\theta ;(z^{\prime}, y))\right\|_{2}^{2} =\|(1-\alpha)\theta_{f, i}^{\top} + \frac{\alpha}{(k-1)}\sum_{j\neq i}^{k}\theta_{f, j}^{\top}-\sum_{j=1}^{k} p_{j}\left(\theta ; z^{\prime}\right) \cdot \theta_{f, j}^{\top}\|_{2}^{2}
\end{equation*}
\par Then, we obtain
\begin{equation*}
\left\|\nabla_{z^{\prime}} \ell\left(\theta ;\left(z, y\right)\right)-\nabla_{z^{\prime}} \ell(\theta ;(z^{\prime}, y))\right\|_{2}^{2}=\left\|\sum_{j=1}^{k}\left(p_{j}(\theta ; z)-p_{j}\left(\theta ; z^{\prime}\right)\right) \theta_{f,j}^{\top}\right\|_{2}^{2}
\end{equation*}
\par We derive $z^\prime$ in $p_j(\theta; z^\prime)$ and get
\begin{equation*}
\begin{aligned}
\left\|\nabla_{z^{\prime}} p_{j}(\theta ; z^{\prime})\right\|^{2}_{2}&
=\left\|-p_{j}(\theta ; z^{\prime})\left(\theta_{f, i}^{\top}-\sum_{j=1}^{k} p_{j}(\theta ; z^{\prime}) \cdot \theta_{f, j}^{\top}\right)\right\|_{2}^{2}\\
& \leq \left\|\theta_{f, i}^{\top}-\sum_{j=1}^{k} p_{j}(\theta ; z^{\prime}) \cdot \theta_{f, j}^{\top}\right\|_{2}^{2}\\
& \leq  2 \max _{1 \leq j \leq k}\left\|\theta_{f, j}\right\|_2^2
\end{aligned}
\end{equation*}
\par Thus, we have
\begin{equation}
\left\|\nabla_{z^{\prime}} \ell\left(\theta ;\left(z, y\right)\right)-\nabla_{z^{\prime}} \ell(\theta ;(z^{\prime}, y))\right\|_{2}\leq 2\sum_{j=1}^{k}\|\theta_{f,j}\|_2\max_{1 \leq j^{'} \leq k}\|\theta_{f, j^{'}}\|_2\left(\|z-z^{\prime}\|_{2}\right)
\label{a21}
\tag{21}
\end{equation}
\subsection{Proof of Theorem \ref{t8}}\label{App6}\
Note that we perform gradient steps with $g^{k} = \nabla_{\theta} g\left(\theta^{k}, z^{k} ; z^{\prime k}\right) = \nabla_{\theta}\boldsymbol\ell\left(\theta ;\left(z^{\prime\star k}, y\right)\right)$. Let $\theta^{k+1}=\theta^{k}-\beta_k g^{k}$. According to Assumption \ref{l4}, we use the $L_{\theta}$-smoothness Taylor expansion for the objective $F$ and get
\begin{equation*}
\begin{aligned}
F(\theta^{k+1}) & \leq F(\theta^{k}) + \nabla F(\theta^{k})(\theta^{k+1}-\theta^{k}) + \frac {L_{\theta}}{2}\left\|\theta^{k+1}-\theta^{k}\right\|_{2}^{2}\\
& = F(\theta^{k}) - \beta_k\nabla F(\theta^{k})g^{k} + \frac {L_{\theta}\beta_k ^{2}}{2}\left\|g^{k}\right\|_{2}^{2}\\
& = F(\theta^{k})+\frac{L_{\theta}\beta_k^{2}}{2}\left\|g^{k}\right\|_{2}^{2}-\beta_k\|\nabla F(\theta^{k})\|_{2}^{2}+\beta_k\nabla F(\theta^{k})(\nabla F(\theta^{k})-g^{k})\\
& = F(\theta^{k})+\frac{L_{\boldsymbol{\phi}}\beta_k^{2}}{2}\left\|g^{k}\right\|_{2}^{2}-\beta_k\|\nabla F(\theta^{k})\|_{2}^{2} +\beta_k(1-L_{\theta}\beta_k)\nabla F(\theta^{k})(\nabla F(\theta^{k})-g^{k}) \\ & + L_{\theta}\beta_k^{2}\nabla F(\theta^{k})(\nabla F(\theta^{k})-g^{k})\\
& = F(\theta^{k})-\beta_k(1-\frac{L_{\theta}\beta_k}{2})\|\nabla F(\theta^{k})\|_{2}^{2}+\beta_k(1-L_{\theta}\beta_k)\nabla F(\theta^{k})(\nabla F(\theta^{k})-g^{k})\\ 
&+\frac{L_{\theta}\beta_k^{2}}{2}\left\|g^{k}-\nabla F(\theta^{k})\right\|_{2}^{2}
\label{a25}
\end{aligned}
\end{equation*}
\par We define the biased error $\delta^{k}=g^{k}-\nabla_{\theta}\boldsymbol\ell(\theta^{k} ;(z^{\prime k}, y))$, then, we get
\begin{equation*}
\begin{aligned}
F(\theta^{k+1}) \leq & F(\theta^{k})-\beta_k(1-\frac{L_{\theta}\beta_k}{2})\left\|\nabla F(\theta^{k})\right\|_{2}^{2}+\beta_k(1-L_{\theta}\beta_k)\nabla F(\theta^{k})(\nabla F(\theta^{k})-\nabla_{\theta}\boldsymbol\ell(\theta^{k} ;(z^{\prime k}, y)))\\
&-\beta_k(1-L_{\theta}\beta_k)\nabla F(\theta^{k})\delta^{k} + \frac{L_{\theta}\beta_k^{2}}{2}\left\|\nabla_{\theta}\boldsymbol\ell(\theta^{k} ;(z^{\prime k}, y))+\delta^{k}-\nabla F(\theta^{k})\right\|_{2}^{2}\\
&\leq F(\theta^{k})-\beta_k(1-\frac{L_{\theta}\beta_k}{2})\left\|\nabla F(\theta^{k})\right\|_{2}^{2}+\beta_k(1-L_{\theta}\beta_k)\nabla F(\theta^{k})(\nabla F(\theta^{k}) - \nabla_{\theta}\boldsymbol\ell(\theta^{k} ;(z^{\prime k}, y))\\
&-\beta_k\left(1-L_{\theta}\beta_k\right)\nabla F(\theta^{k})\delta^{k}+\frac{L_{\theta}\beta_k^{2}}{2}
\left(\left\|\nabla_{\theta}\boldsymbol\ell(\theta^{k} ;(z^{\prime k}, y))-\nabla F(\theta^{k})\right\|_{2}^{2} \right. \\ 
&+\left. \left\|\delta^{k}\right\|_{2}^{2} + 2 (\nabla_{\theta}\boldsymbol\ell(\theta^{k} ;(z^{\prime k}, y))-\nabla F(\theta^{k}))\delta^{k}\right)
\end{aligned}
\end{equation*}
\par Using inequality $ 2ab \leq \|a\|_{2}^{2} + \|b\|_{2}^{2}$, we obtain
\begin{equation}
\begin{aligned}
F(\theta^{k+1}) \leq & F(\theta^{k})-\frac{\beta_k}{2}\left\|\nabla F(\theta^{k})\right\|_{2}^{2} +\beta_k(1-L_{\theta}\beta_k)\nabla F(\theta^{k})(\nabla F(\theta^{k})-\nabla_{\theta}\boldsymbol\ell(\theta^{k} ;(z^{\prime k}, y))) \\ &
+\frac{\beta_k(1+L_{\theta}\beta_k)}{2}\left\|\delta^{k}\right\|_{2}^{2} + L_{\theta}\beta_k^{2} \left\|\nabla_{\theta}\boldsymbol\ell(\theta^{k} ;(z^{\prime k}, y))-\nabla F(\theta^{k}) \right\|_{2}^{2}
\end{aligned}
\tag{22}
\end{equation}
\par Using Assumption \ref{l4} and Theorem \ref{t6}, the error $\delta^{k}$ satisfies
\begin{equation}
\begin{aligned}
\left\|\delta^{k}\right\|_{2}^{2}&=\left\|\nabla_{\theta} \boldsymbol\ell(\theta^{k} ; (z^{\prime\star k},y))-\nabla_{\theta}\boldsymbol\ell(\theta^k ;\left(z^{\prime k}, y\right))\right\|_{2}^{2}\\ 
&\leq L_1^2\left\|z^{\prime\star k} - z^{\prime k}\right\|_{2}^{2}\leq \frac{3L_0L_1^2}{\gamma}
\end{aligned}
\label{a22}
\tag{23}
\end{equation}
\par Using bound (\ref{a22}), $ E[\left\|\nabla F(\theta)-\nabla_{\theta} \boldsymbol{\phi_{\gamma}}(\theta, (z,y))\right\|_{2}^{2}]\leq\sigma^{2}$, and ${E}[\nabla_{\theta}\boldsymbol{\phi_{\gamma}}\left(\theta^{k} ; (z^{k},y)\right) \mid \theta^{k}]= \nabla F \left(\theta^{k}\right)$, we obtain
\begin{equation*}
\begin{aligned}
{E}[F(\theta^{k+1})-F(\theta^{k})\mid \theta^{k}]&\leq -\frac{\beta_k}{2}\left\|\nabla F(\theta^{k})\right\|_{2}^{2}+\frac{3\beta_k(1+L_{\theta}\beta_k)L_0L_1^2}{2\gamma} + L_{\theta}\beta_k^{2}\sigma^{2}
\end{aligned}
\end{equation*}
\par Since $\beta_k \leq \frac{1}{L_\theta}$, we have
\begin{equation*}
\begin{aligned}
\frac{\beta_k}{2}\left\|\nabla F(\theta^{k})\right\|_{2}^{2} &\leq E\left[F(\theta^{k})-F\left(\theta^{k+1}\right)\right]+\frac{3\beta_k(1+L_{\theta}\beta_k)L_0L_1^2}{2\gamma} + L_{\theta}\beta_k^{2}\sigma^{2}\\
& \leq E\left[F(\theta^{k})-F\left(\theta^{k+1}\right)\right]+\frac{3\beta_kL_0L_1^2}{\gamma} + L_{\theta}\beta_k^{2}\sigma^{2}
\end{aligned}
\end{equation*}
For a fixed constant step size $\beta$, we obtain
\begin{equation*}
\begin{aligned}
\left\|\nabla F(\theta^{k})\right\|_{2}^{2} &\leq \frac{2}{\beta} E\left[F(\theta^{k})-F\left(\theta^{k+1}\right)\right]+\frac{3L_0L_1^2}{\gamma} + 2L_{\theta}\beta\sigma^{2}
\end{aligned}
\end{equation*}
\par Summing over $k$ times and using $\Delta_{F} \geq F\left(\theta^{1}\right)-\inf _{\theta} F(\theta)$, we arrive
\begin{equation}
\begin{aligned}
\frac{1}{K}\sum_{k=1}^{K} {E}\left[\left\|\nabla F(\theta^{k})\right\|_{2}^{2}\right] \leq &\frac{2}{\beta K} E\left[F\left(\theta^{1}\right)-F(\theta^{K})\right] + \frac{3L_0L_1^2}{\gamma} + 2L_{\theta}\beta\sigma^{2}\\
&\leq \frac{2\Delta_{F}}{\beta K}  + \frac{3L_0L_1^2}{\gamma} + 2L_{\theta}\beta\sigma^{2}
\end{aligned}
\label{a23}
\tag{24}
\end{equation}
\par We set $\beta = \sqrt{\frac{2\Delta_{F}}{L_{\theta}K\sigma^2}}$ to get the result.
\end{APPENDIX}
\end{document}